\documentclass[letterpaper, 10 pt, conference]{ieeeconf}  %

\IEEEoverridecommandlockouts                              %

\usepackage[table]{xcolor} 
\usepackage{tabularx}
\usepackage{multirow}
\usepackage{cite}
\usepackage{caption}
\usepackage{amsmath,amssymb,amsfonts}
\usepackage{algorithmic}
\usepackage{graphicx}
\usepackage{textcomp}
\usepackage{booktabs} %
\usepackage{hhline}
\usepackage{arydshln}
\usepackage[pagebackref=true,breaklinks=true,letterpaper=true,colorlinks,bookmarks=true]{hyperref}
\usepackage{fancyhdr}

\newcommand{\model}{DISORF}

\title{\LARGE \bf
\model: A Distributed Online 3D Reconstruction Framework for Mobile Robots
\thanks{*Authors contributed equally.}
}

\author{Chunlin Li$^{1*}$, Hanrui Fan$^{1*}$, Xiaorui Huang$^{1*}$, Ruofan Liang$^{2}$, Sankeerth Durvasula$^{2}$, Nandita Vijaykumar$^{2}$
\thanks{Department of Computer Science, University of Toronto, Canada}
\thanks{$^{1}${\footnotesize \{\href{mailto:edwardphi.li@mail.utoronto.ca}{edwardphi.li}, \href{mailto:hanrui.fan@mail.utoronto.ca}{hanrui.fan}, \href{mailto:richardxr.huang@mail.utoronto.ca}{richardxr.huang}\}@mail.utoronto.ca}}
\thanks{$^{2}${\footnotesize \{\href{mailto:ruofan@cs.toronto.edu}{ruofan}, \href{mailto:sankeerth@cs.toronto.edu}{sankeerth}, \href{mailto:nandita@cs.toronto.edu}{nandita}\}@cs.toronto.edu}}
\thanks{Digital Object Identifier (DOI): see top of this page.}
}

\begin{document}

\maketitle

\pagestyle{fancy}
\fancyhf{}
\fancyfoot[L]{\footnotesize This work has been submitted to the IEEE Robotics and Automation Letters for possible publication. Copyright may be transferred without notice, after which this version may no longer be accessible.}
\renewcommand{\headrulewidth}{0pt}

\begin{abstract}
We present a framework, \model, to enable online 3D reconstruction and visualization of scenes captured by resource-constrained mobile robots and edge devices. To address the limited computing capabilities of edge devices and potentially limited network availability, we design a framework that efficiently distributes computation between the edge device and the remote server.   
We leverage on-device SLAM systems to generate posed keyframes and transmit them to remote servers that can perform high-quality 3D reconstruction and visualization at runtime by leveraging recent advances in neural 3D methods.
We identify a key challenge with online training where naive image sampling strategies can lead to significant degradation in rendering quality.
We propose a novel shifted exponential frame sampling method that addresses this challenge for online training. 
We demonstrate the effectiveness of our framework in enabling high-quality real-time reconstruction and visualization of unknown scenes as they are captured and streamed from cameras in mobile robots and edge devices.

\end{abstract}


\section{Introduction}
\label{sec:intro}
Online 3D reconstruction to learn a representation of a scene at real-time{\textemdash}where RGB images are continuously captured and used to optimize a 3D model that can be rendered and visualized{\textemdash}holds immense potential in various domains. 
For example, mobile robots and embodied devices (e.g., drones) navigating a previously unseen environment with the ability to construct and visualize a 3D model of the environment \emph{on the fly}, offer opportunities for enhanced navigation, scene understanding, and interactive exploration of the environment. 
Online 3D scene reconstruction has been extensively studied with various methods to represent geometry and appearance based on voxel or point representations~\cite{voxel_hashing, bundlefusion, whelan2016elasticfusion}. 
Recently, implicit neural representation methods such as neural radiance fields (NeRFs)~\cite{mildenhall2020nerf} and 3D Gaussian Splatting (3DGS)~\cite{kerbl20233d} have emerged as promising approaches to represent complex 3D scenes with the capability of photorealistic 3D scene rendering and visualization.

NeRF methods use multi-view posed images to learn a neural implicit function (usually an MLP) optimized through differentiable volumetric rendering. 
3DGS learns and explicitly represents the scene with 3D Gaussians, leveraging differentiable rasterization and alpha-blending to achieve interactive rendering speed.
NeRF/3DGS is mostly used as an offline 3D scene representation method as its training process requires sampling over multi-view images with well-calibrated camera parameters.  
However, to make these approaches compatible with an incremental online learning process, the camera poses of captured frames must be estimated on the fly. This can be done by leveraging the tracking module in real-time simultaneous localization and mapping (SLAM) systems commonly used in mobile robots. 
Recent research has also investigated the potential of leveraging SLAM for online 3D reconstruction.
However, several challenges must be addressed to make neural online 3D scene representation feasible in mobile robots. 

\begin{figure}
    \centering
    \includegraphics[width=\linewidth, trim={0 0pt 0pt 0pt},clip]{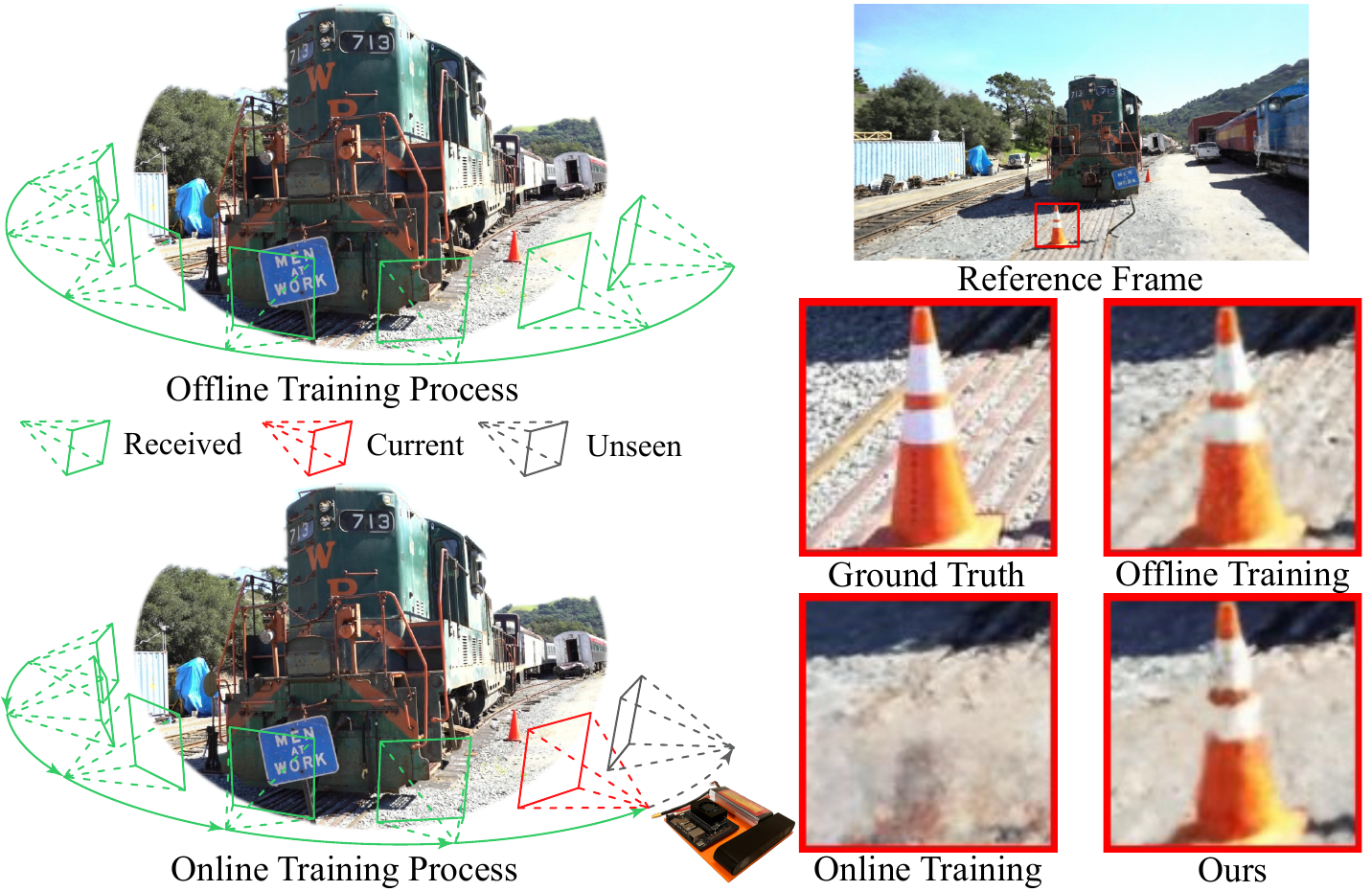}
    \caption{The setup and rendering results for online and offline training: For offline training  (top), images from all viewpoints are available throughout the training process. For online training (bottom), the model is continuously trained as new images are streamed.
    }
    \label{fig:offline_online}
    \vspace{-8mm}
\end{figure}

First, mobile and embodied robots or edge devices are resource constrained. This makes it difficult to support powerful GPUs and large memory capacities. Thus, performing computationally expensive NeRF/3DGS training on resource-constrained edge devices is usually impractical. 
Although recent advances have seen great improvements in accelerating training and rendering speed \cite{muller2022instant, kerbl20233d},
the substantial computing requirement for efficient training cannot be satisfied on resource-constrained edge computing platforms. For example, a powerful edge GPU device like Jetson Xavier NX still takes over 14 times longer~\cite{instant3d} than an RTX3090 GPU to train Instant-NGP, currently one of the most efficient NeRF models.
Devices with even less computing power, such as Jetson TX2 and Raspberry Pi would incur significantly larger training latencies.
Distributing the intense compute requirement to a \emph{remote server} that is provisioned with powerful compute resources is a promising approach to enable online NeRF/3DGS training and rendering. 
Recent work~\cite{yu2023nerfbridge} has developed a framework for transmitting image streams from edge cameras and optimizing them offline on a remote server. However, this approach relies \emph{solely} on remote computation, neglecting the potential benefits of using mobile robots' visual odometry and localization capabilities for pose estimation. Additionally, it assumes a constant and sufficient network bandwidth, limiting its applicability to specific domains and use cases.

\begin{figure}
    \centering
    \includegraphics[width=\linewidth, trim={15pt 15pt 15pt 15pt},clip]{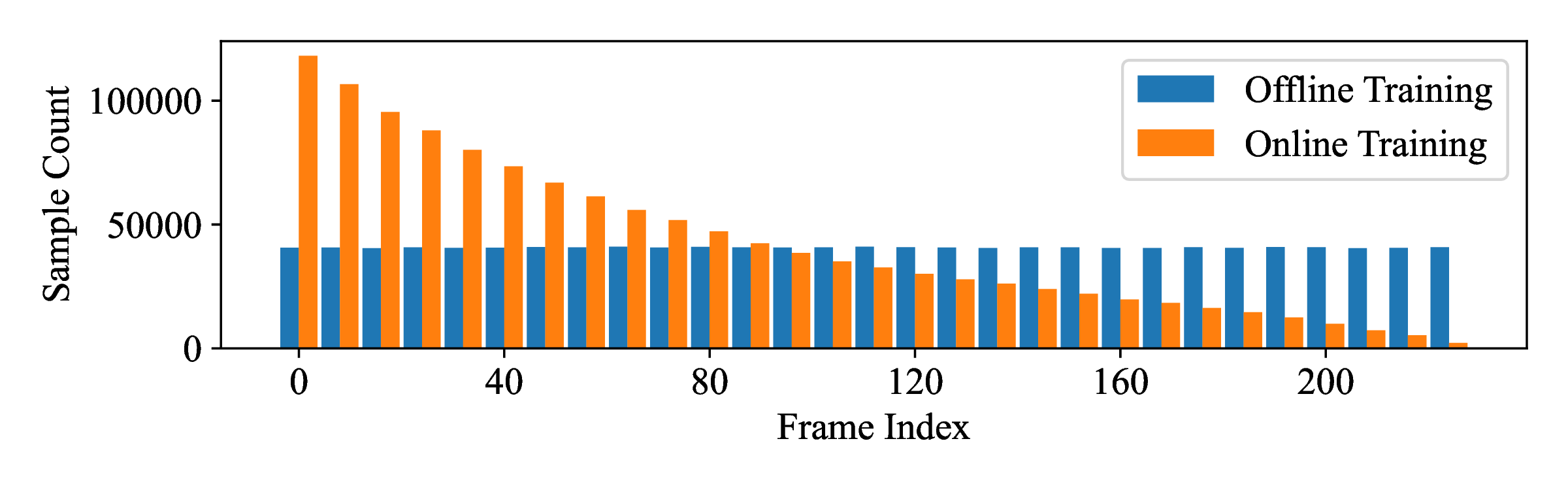}
    \vspace{-18pt}
    \caption{The total number of pixels from each frame being sampled after NeRF training on a Replica one-minute keyframe stream.}
    \label{fig:sample_count}
    \vspace{-6mm}
\end{figure}

Second, a critical challenge in enabling high-quality online 3D reconstruction is the sampling strategy for online training. NeRF/3DGS training requires sampling pixels/frames from captured images of the represented scene. The most commonly used sampling method is random uniform sampling, which uniformly samples $N$ rays/images from all currently existing images for each training iteration. This approach is effective for offline training, but results in sub-optimal rendering with online training as shown in Fig. \ref{fig:offline_online}.
To illustrate this challenge, let's consider an example comparing online and offline NeRF training scenarios. In both cases, we train NeRF/3DGS models using keyframes output by on-device ORB-SLAM2~\cite{mur2017orb}. 
This quality drop can be explained by the imbalance in frame sampling distribution in the online scenario. As seen in Fig. \ref{fig:sample_count}, offline training samples a roughly equal number of pixels from each training frame, whereas online training samples more from the earlier frames and less from the recent frames overall. This is the case because images are continuously streamed during training, thus earlier images will be sampled in more iterations. Since the more recent frames are less sampled, objects in these frames may not be sufficiently trained in the online training process to optimize their shape and appearance, leading to lower rendering quality.

We aim to enable online 3D reconstruction and visualization of environments/scenes from mobile robots and edge devices by addressing the aforementioned challenges.
To achieve this, we introduce \model, a novel framework that enables online 3D reconstruction with NeRF and 3DGS by distributing the computational tasks between the edge devices and a remote server. With \model, we leverage on-device SLAM for pose estimation and only transmit keyframes to the remote server for processing. This approach effectively reduces the reliance on network bandwidth for high-quality reconstruction. The resource-intensive NeRF/3DGS computations are performed on powerful servers, which are essential for online training and rendering.
 
We explore input sampling strategies for online 3D reconstruction and find that giving greater weight to recent frames during each iteration visibly enhances reconstruction quality. Building on this insight, we introduce a shifted exponential sampling weight function. This function dynamically focuses on recently received frames during training, mitigating the issue of inadequate samples for recent frames.

\section{Related Work}

\subsection{Online 3D Representation}
\label{sec:classical_online_reconstruction}

Unlike offline reconstruction methods that are able to access all the frames for iterative global optimizations \cite{schoenberger2016sfm}, the online 3D reconstruction task requires on-the-fly camera pose estimation and reconstruction. Therefore, most online 3D reconstruction is closely integrated with SLAM systems.
These dense 3D reconstruction methods~\cite{6162880, voxel_hashing} typically use RGB-Depth sensors as input to incrementally reconstruct scene geometry using either the signed distance field (SDF) or occupancy estimated with voxels as 3D scene representations and some approaches use explicit representations such as surfels~\cite{keller2013real, cao2018real}. 
With the emergence of neural representations, recent works have leveraged neural networks for online 3D reconstruction~\cite{sucar2020nodeslam, weder2021neuralfusion}.
Recent neural reconstruction methods also enable promising online 3D reconstruction results from monocular video 
streams~\cite{sun2021neuralrecon, teed2021droid}. 
These prior works aim to enable high-quality 3D surface or geometry reconstruction and do not address the challenges of enabling high-quality online 3D reconstruction and visualization on resource-constrained mobile robots and edge devices.

\subsection{Neural 3D Representations and Robotics Applications}
3D representations have significantly evolved with the success of differentiable optimizations. 
NeRF \cite{mildenhall2020nerf}, a pivotal work in this area, is an implicit neural scene representation that enables photorealistic view synthesis given a set of multiview posed images.
The key idea is to use a neural network (e.g., an MLP) to represent implicit fields (e.g., volume density and color), which are used for volumetric rendering and optimized using a photometric loss.
The initial NeRF architecture is compute-intensive, requiring hours or even days of training, while incurring high rendering latencies.
However, recent advances \cite{muller2022instant, chen2022tensorf} have significantly accelerated NeRF's training and rendering speed, reducing training time to minutes or even seconds.
3DGS \cite{kerbl20233d}, another novel 3D representation, has recently attracted great attention in the research community. It represents 3D scenes using discrete 3D Gaussian points, with differentiable rasterization and alpha-blending to render these 3D Gaussians, significantly improving rendering efficiency.

The promising capabilities of these neural 3D methods have also been leveraged for various robotics tasks.
Evo-NeRF \cite{kerr2022evo} trains NeRFs online while grasping using a robotic arm. This is primarily focused on the high-speed training of NeRFs given an evolving scene as opposed to achieving high-quality online 3D reconstruction. Existing works \cite{yen2022nerf, byravan2023nerf2real, adamkiewicz2022vision, lu2024manigaussian, wang2024query} also leverage NeRF/3DGS to facilitate downstream tasks such as trajectory planning, training control policies for robot motion and manipulation, and more. However, these works mainly use a pre-trained 3D model for various downstream tasks, while our work focuses on online training. Our framework could potentially be used in these scenarios to enable real-time applications.

\subsection{Incremental Learning for 3D}

Incremental or continual learning involves a machine learning model progressively learning new knowledge from a training dataset that continually expands as new data is gathered over time. 
Some existing works leverage the idea of continual learning to represent temporally evolving scenes with NeRF \cite{cai2023clnerf}.
Online NeRF/3DGS training can also be treated as a replay-based incremental learning approach~\cite{rebuffi2017icarl}.
There are existing works that perform incremental training in SLAM systems for dense reconstruction with NeRF~\cite{sucar2021imap, zhu2022nice} or 3DGS~\cite{matsuki2024gaussian, keetha2024splatam, huang2024photo}. 
These neural SLAM methods often manually pre-define a fixed portion of training samples from recent frames and the remainder from earlier frames in each training iteration.
In contrast, our sampling method offers a smoother transition from concentrated to uniform sampling as the training iteration progresses and the number of training frames increases. This enhances its adaptability across diverse scenes and scenarios, ensuring balanced and effective learning throughout the training process. However, these existing methods are still costly to deploy or run resource-constrained edge devices.

\section{Framework Design}

\begin{figure}
    \vspace{1mm}
    \centering
    \includegraphics[width=\linewidth, trim={0 0pt 0pt 0pt},clip]{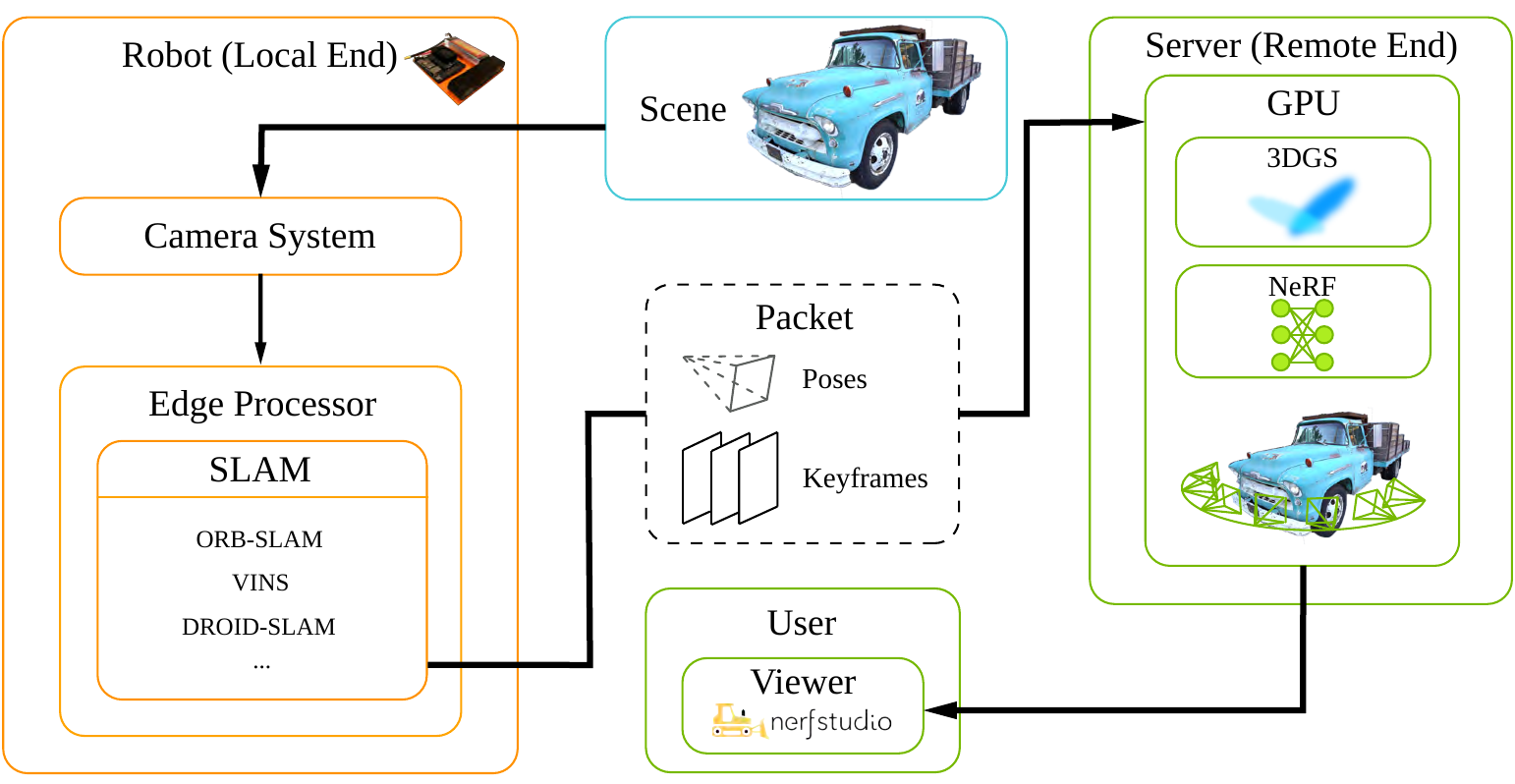}
    \caption{\model{} pipeline distributing computation between a mobile robot and a remote server. The robot's camera captures images processed by the edge processor's SLAM module. Keyframes and poses are streamed to the server, where GPU-intensive 3D model training (NeRF/3DGS) is performed.}
    \label{fig:process_chart}
    \vspace{-0.5cm}
\end{figure}

We introduce our proposed framework, \model, designed for online 3D scene reconstruction and visualization using images streamed from a mobile robot or edge device, as illustrated in Fig. \ref{fig:process_chart}. This section details the key modular components of our framework, including the SLAM module running on the local robot, the 3D reconstruction module running on the remote server, and the network connections between them.

\subsection{SLAM Module on Local Robot}

Our framework's local end is deployed on a robot's embedded computer. Its primary function is to determine the robot's poses in real-time using SLAM algorithms and prepare posed images for the server to reconstruct the scene. Numerous off-the-shelf SLAM algorithms are available, each with unique features \cite{mur2017orb, qin2018vins, teed2021droid}.
Our framework is designed to integrate open-source SLAM algorithms into the local end, while ensuring compatibility with the rest of our system.
For robustness and generalizability, we use ORB-SLAM2 \cite{mur2017orb} as the representative SLAM algorithm for our evaluations. Depending on the sensors and computing capabilities of the robot, other SLAM systems can be used. For example, VINS-Mono \cite{qin2018vins} if an IMU sensor is available; RTAB-Map \cite{labbe2019rtab} if a Lidar sensor is available; and Droid-SLAM \cite{teed2021droid} if the robot has a powerful GPU.
The output of our local end consists of posed keyframes estimated by the tracking module of the deployed SLAM system. Those keyframes, along with related attributes such as camera poses and timestamps, are then transmitted to the remote server for the online neural reconstruction.

\subsection{3D Neural Reconstruction on Remote Server}

In our framework, the 3D reconstruction is performed on a remote server equipped with sufficient computational resources to run NeRF and 3DGS training. The server receives posed key frames from the local robot via the network and stores them in a database. When receiving this data, the server initializes the appropriate training and rendering modules based on the pre-defined configuration. It then processes the data for online training and rendering tasks in real-time.
Our server's software stack is built on a modified version of Nerfstudio \cite{tancik2023nerfstudio}. This adapted Nerfstudio is utilized for training, visualizing, and rendering the reconstruction results when the robot explores the scene. By leveraging the computational power of the remote server, this approach ensures efficient, robust, and high-quality 3D reconstruction. We detail the model and training adaptations in Sec. \ref{sec:model} and Sec. \ref{sec:sampling}.

\subsection{Network Connection for Distributed Computing}
We use standard transport protocols from ROS \cite{quigley2009ros} for real-time data transmission from the robot's local end to the server's remote end over the network. On the local end, once the SLAM module outputs the estimation of the next keyframe, the publisher then sends this frame data packet, consisting of the image with its pose and timestamp. The remote end, via its subscriber, receives the frame data and stores it in the server for model training.

In our distributed computing workflow, we transmit only the keyframe data obtained from the local SLAM module\footnote{Depending on the SLAM system, keyframes can be determined by picture/camera motions or simply pre-defined frame sub-sampling.} instead of all frames from the video stream. The typical keyframe rate is 2$\sim$5 FPS, much lower than the full video frame rate, which is usually over 20 FPS. Compared to previous solutions that transmit all frames \cite{yu2023nerfbridge}, our approach significantly reduces network bandwidth requirements, making it more resilient to varying network conditions. For example, our keyframe transmission reduces network bandwidth usage by nearly 9x on Replica scans (0.91 Mbps vs 8.14 Mbps).

\section{Online 3D Neural Reconstruction Models} \label{sec:model}
Our framework is designed to be model-agnostic, allowing users to choose the neural scene reconstruction model that best fits their needs. For instance, with limited training time and no sparse 3D geometry prior, the NeRF-based model \cite{barron2022mip} is well-suited for unbounded outdoor scenes, while Gaussian Splatting \cite{kerbl20233d} excels in real-time rendering framerate and high-frequency detail reconstruction for bounded indoor scenes. 
Our evaluation in Sec. \ref{sec:exp} presents results from both our adapted NeRF and 3DGS models. We will briefly describe our model adaptations below.

\subsection{Neural Radiance Field}
Our NeRF model is based on NeRFacto \cite{tancik2023nerfstudio}, a robust implementation that incorporates many recent advancements in NeRF \cite{barron2022mip, muller2022instant}, making it suitable for general 3D scene reconstruction. To handle the unbounded 3D scenes, NeRFacto uses a non-linear space construction function that maps the entire 3D space into a cubic space bounded within $[-2, 2]$:
\begin{equation}
        f(x) = \begin{cases}
        x & ||x|| \leq 1 \\
        \left(2 - \frac{1}{||x||}\right)\left(\frac{x}{||x||}\right) & ||x|| > 1
    \end{cases}
\end{equation}
where $||\cdot||$ is $L_\infty$ norm. Due to this mapping, the reconstruction quality decreases as the target distance from the origin increases. Thus, offline NeRF training usually involves recentering and rescaling all camera poses before training. However, in our online setup, we do not have access to all camera poses initially. 
To address this, we use the first $k$ camera poses ($k=5$) to estimate the center $o$ of the shared view frustum within a bounded depth region. We then recenter and rescale the current and subsequent camera poses so that $o$ becomes the origin of the new coordinate system, and the first $k$ cameras are placed within the $[-1,1]$ linear cubic space.
Although this approach does not guarantee perfect centering of the region of interest (ROI), it enhances the representation quality by better utilizing the linear space in online NeRF training.

\subsection{3D Gaussian Splatting}
Unlike NeRF, 3DGS explicitly represents the 3D scenes with many discrete Gaussian points, avoiding the issue of space contraction. Optimal 3DGS reconstruction typically relies on some 3D geometry prior (e.g., sparse point cloud from COLMAP \cite{schoenberger2016sfm} of the scene). However, such geometry prior is not available for online training of unseen scenes. 
Therefore, we start 3DGS training with randomly initialized points. We randomly sample $N$ points within [-1,1] cubic space and additionally sample $\frac{1}{2}N$ points within $[-2, 2]$, applying $f^{-1}(x)$ to map points to distant regions. 
To enhance training efficiency and quality, we adopt insights from RAIN-GS \cite{jung2024relaxing}, initializing with a small number of large variance Gaussian points to encourage coarse-to-fine scene reconstruction. We typically set $N$ to 5000 for training. To ensure that newly received frames are well represented with adequate Gaussian points, we continuously perform dynamic Gaussian point pruning and splitting throughout our online training, stopping only after the last keyframe is received.

\subsection{Training Schedulers}
\begin{figure}[h]
    \vspace{0.5mm}
    \centering
    \includegraphics[width=\linewidth]{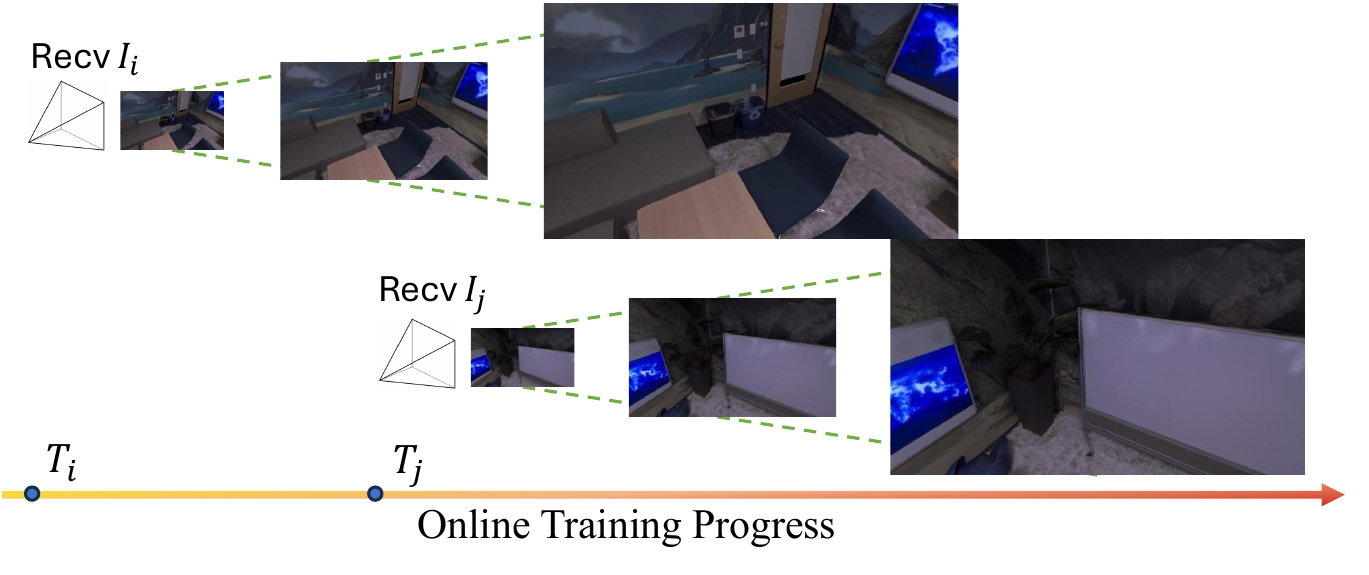}
    \vspace{-20pt}
    \caption{Each received frame is initially trained at a lower resolution, then gradually increased to the original resolution.}
    \label{fig:per-frame}
    \vspace{-4mm}
\end{figure}

In offline NeRF/3DGS training, global schedulers for training loss and learning rate are applied across all frames. However, global training schedulers may not be suitable for online training, as later-received training frames might start with significantly decreased loss or learning rate. To address this, our online NeRF/3DGS training uses a per-frame scheduling strategy, ensuring that newly received frames begin training with initial values and are independently scheduled in subsequent iterations. We use the following scheduling strategies:
\begin{itemize}
    \item \textbf{Learning rate scheduler.} We modify the original global learning rate scheduler to a per-frame scheduler.
    \item \textbf{Training frame resolution.} To facilitate the coarse-to-fine training approach, we initially use downscaled frames, gradually upscaling them to their original resolution through discrete exponential scheduling with a factor of 2 (Fig. \ref{fig:per-frame}). This resolution scheduling strategy has been shown to improve reconstruction quality \cite{huang2024photo}.
\end{itemize}

\section{Online Training Sampling Strategy}\label{sec:sampling}

\begin{table*}[t]
    \vspace{0.6mm}
    \centering
    \footnotesize
    \setlength\tabcolsep{0pt}

    \begin{tabularx}{\linewidth}%
    {p{1em}>{\centering}p{0.265\linewidth} *{3}{>{\centering\arraybackslash}X}}
\rotatebox{90}{\scriptsize $\alpha$=1,$\beta$=0 } 
                     & \includegraphics[width=0.98\linewidth, trim={42pt 88pt 44pt 26pt}, clip]{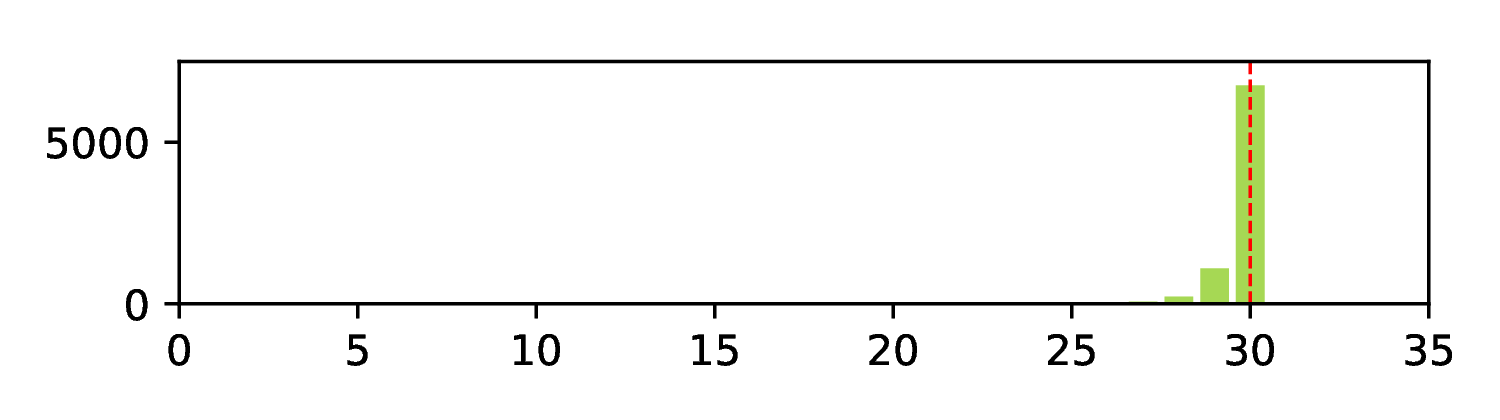} 
                     & \includegraphics[width=0.98\linewidth, trim={168pt 88pt 44pt 26pt}, clip]{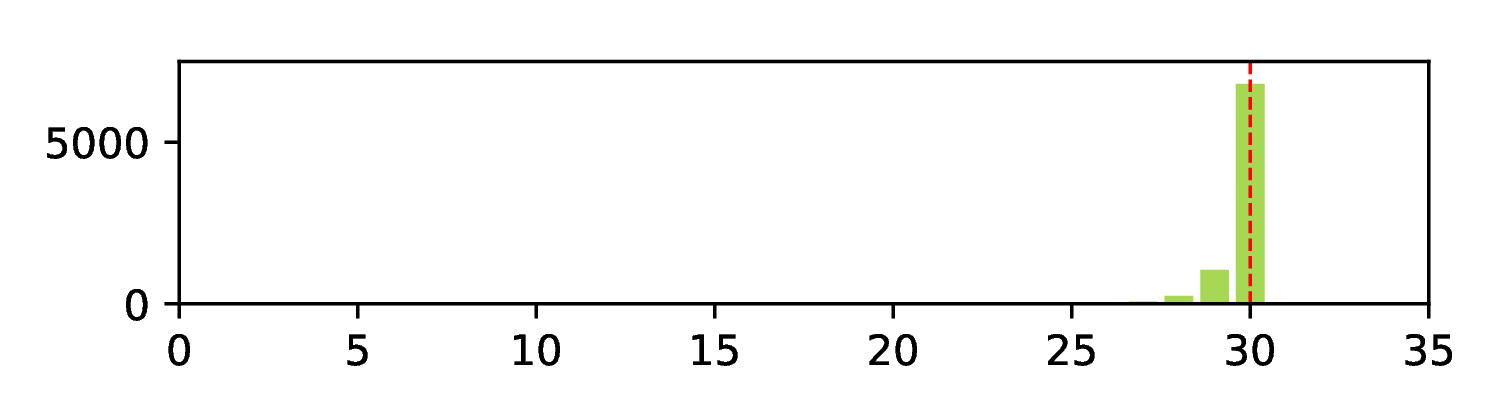}
                     & \includegraphics[width=0.98\linewidth, trim={168pt 88pt 44pt 26pt}, clip]{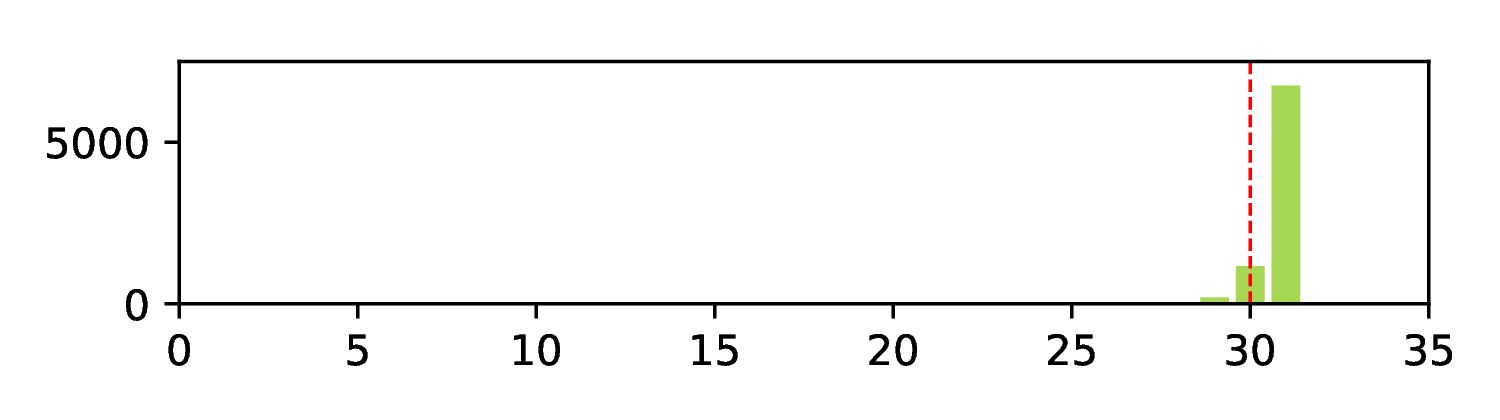} 
                     & \includegraphics[width=0.98\linewidth, trim={168pt 88pt 44pt 26pt}, clip]{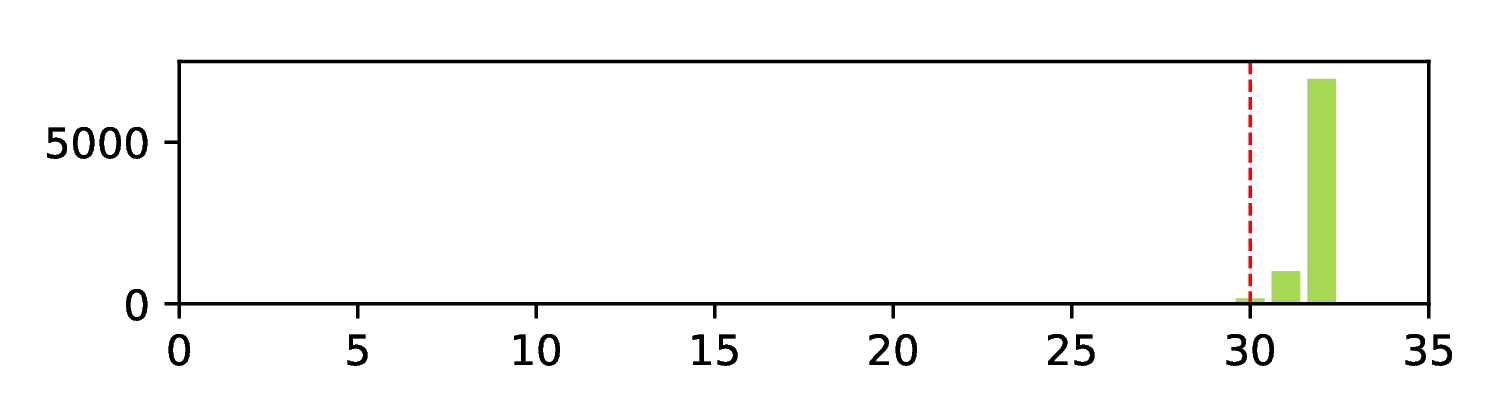}
\\ \hdashline
\rotatebox{90}{\scriptsize $\alpha$=1,$\beta$=8} 
                     & \includegraphics[width=0.98\linewidth, trim={42pt 88pt 44pt 26pt}, clip]{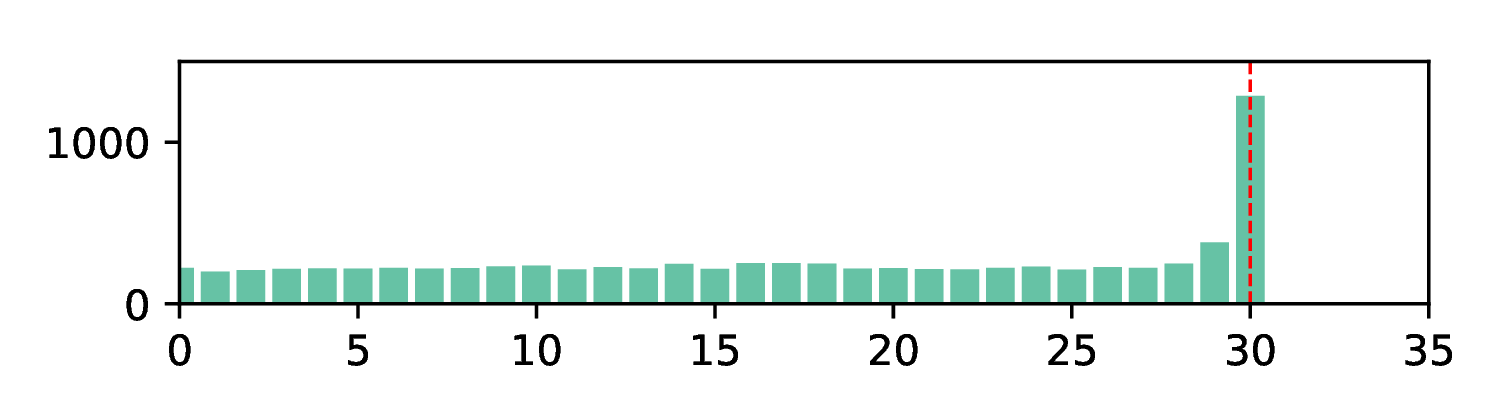} 
                     & \includegraphics[width=0.98\linewidth, trim={168pt 88pt 44pt 26pt}, clip]{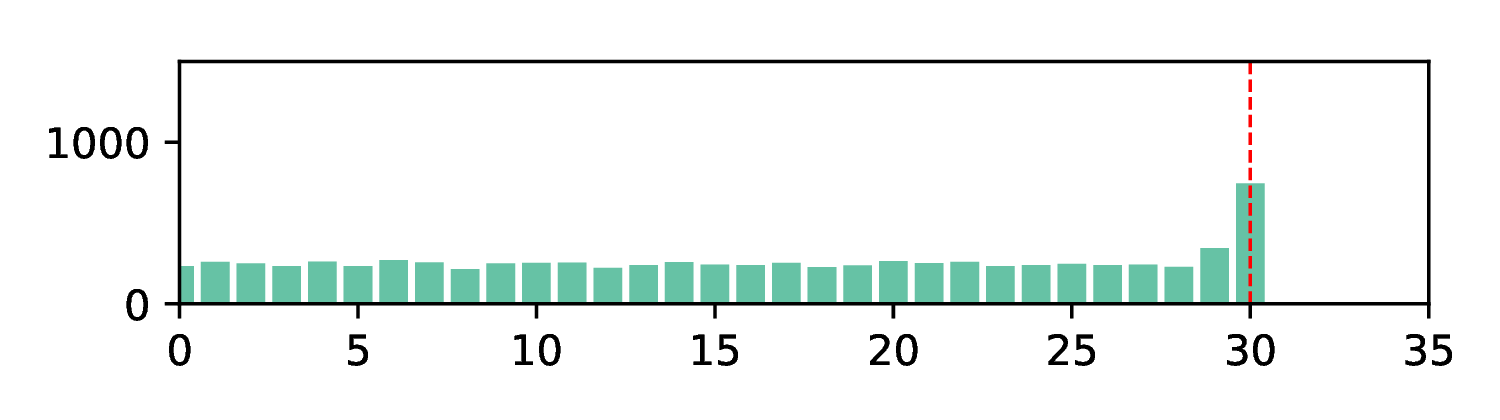}
                     & \includegraphics[width=0.98\linewidth, trim={168pt 88pt 44pt 26pt}, clip]{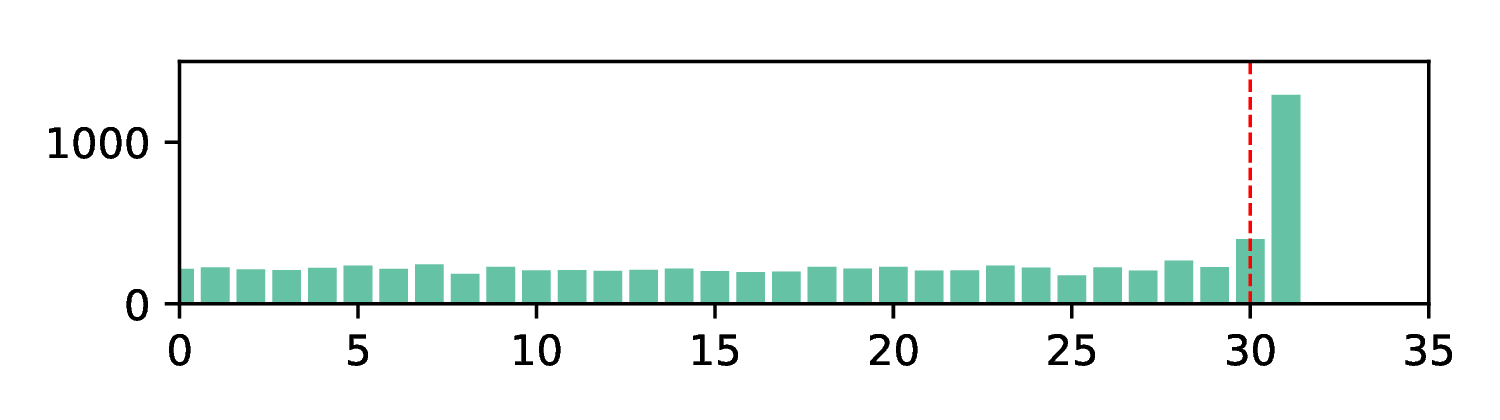} 
                     & \includegraphics[width=0.98\linewidth, trim={168pt 88pt 44pt 26pt}, clip]{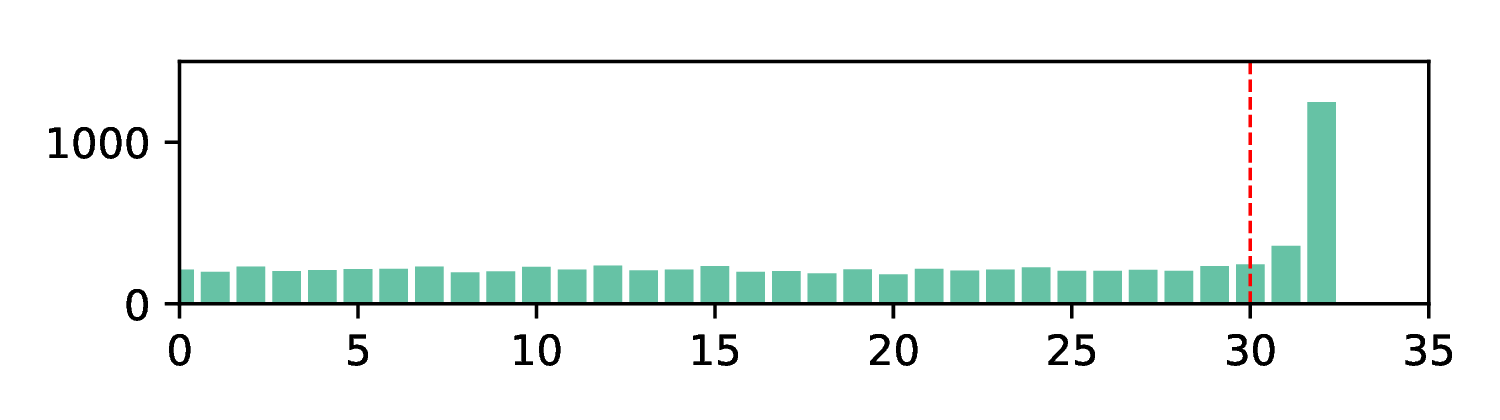}
\\ \hdashline
\rotatebox{90}{\scriptsize $\alpha$=2,$\beta$=8 } 
                     & \includegraphics[width=0.98\linewidth, trim={42pt 88pt 44pt 26pt}, clip]{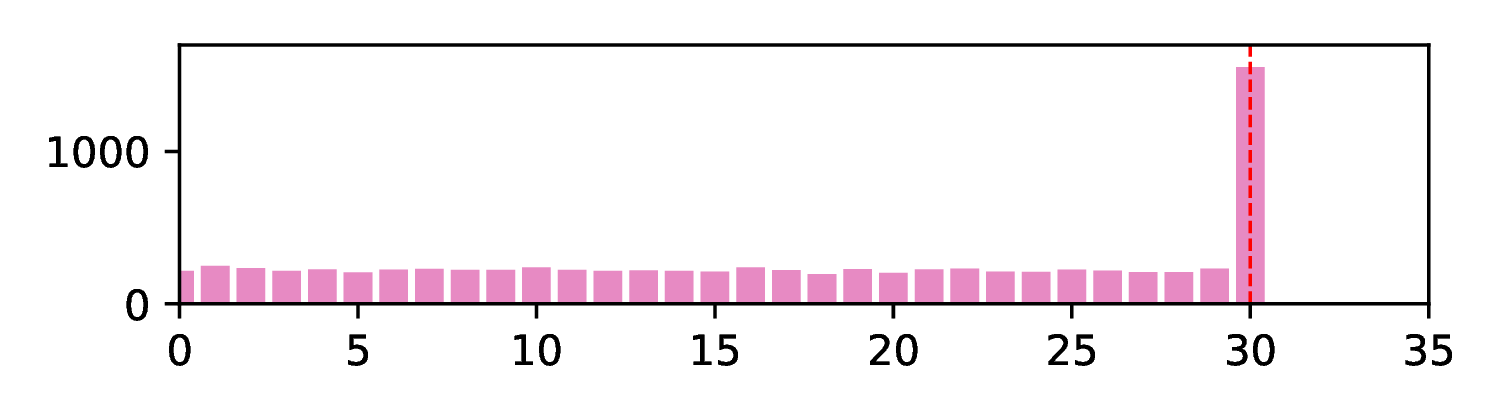} 
                     & \includegraphics[width=0.98\linewidth, trim={168pt 88pt 44pt 26pt}, clip]{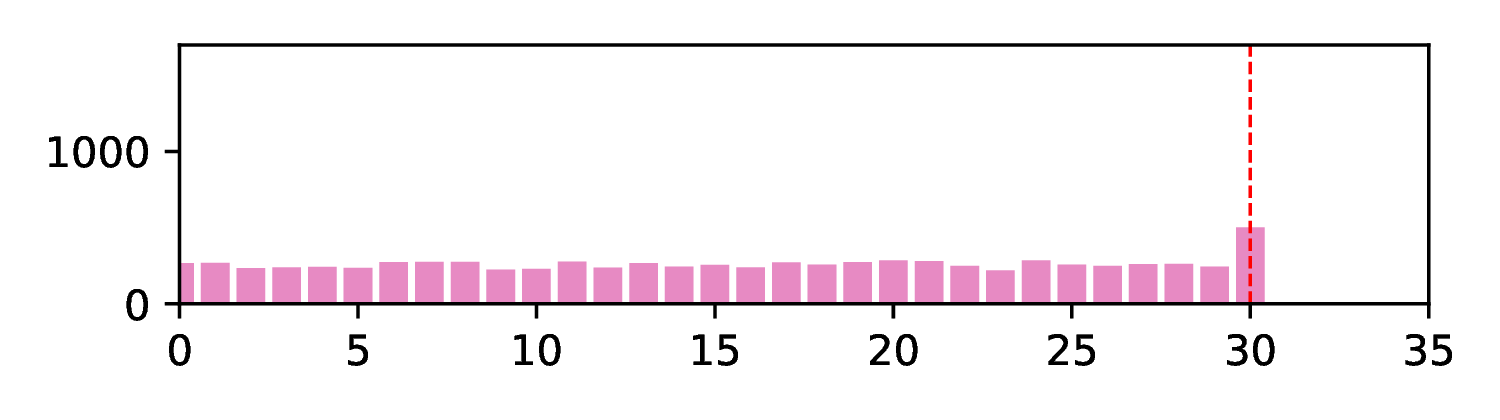}
                     & \includegraphics[width=0.98\linewidth, trim={168pt 88pt 44pt 26pt}, clip]{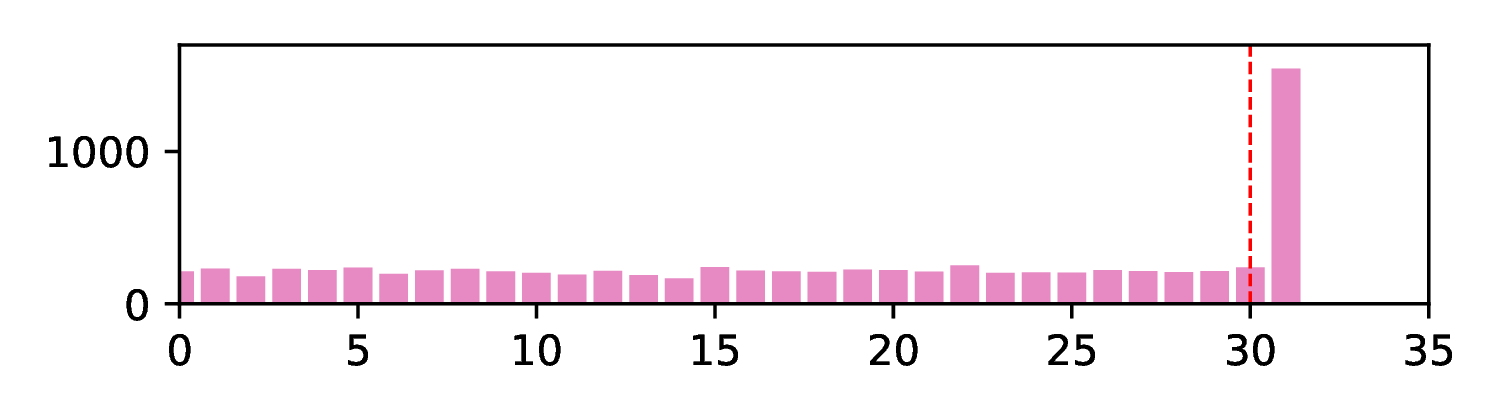} 
                     & \includegraphics[width=0.98\linewidth, trim={168pt 88pt 44pt 26pt}, clip]{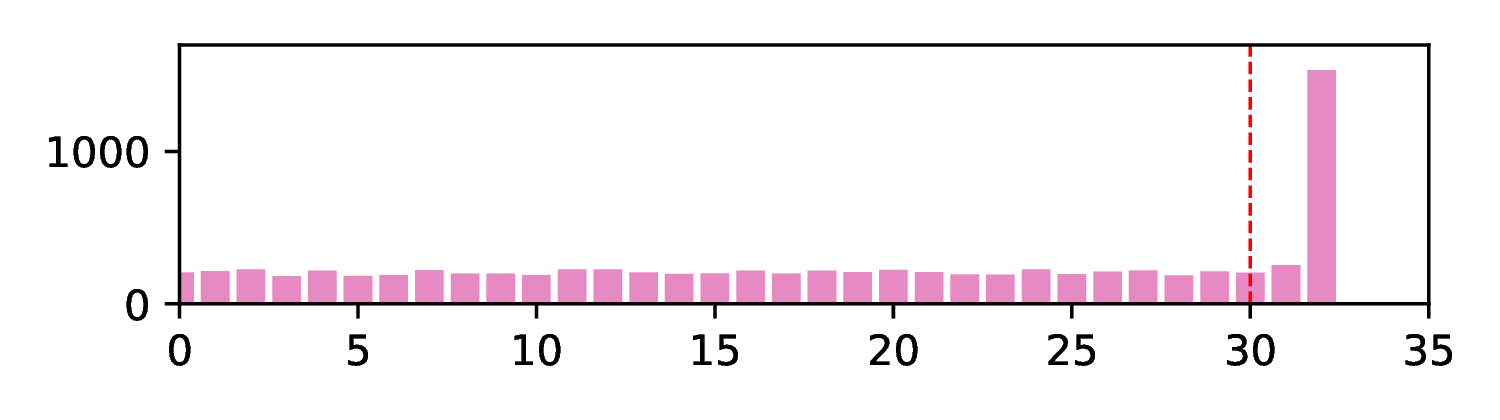}
\\ \hdashline
\rotatebox{90}{\scriptsize $\alpha$=0.5,$\beta$=8 } 
                     & \includegraphics[width=0.98\linewidth, trim={42pt 55pt 44pt 26pt}, clip]{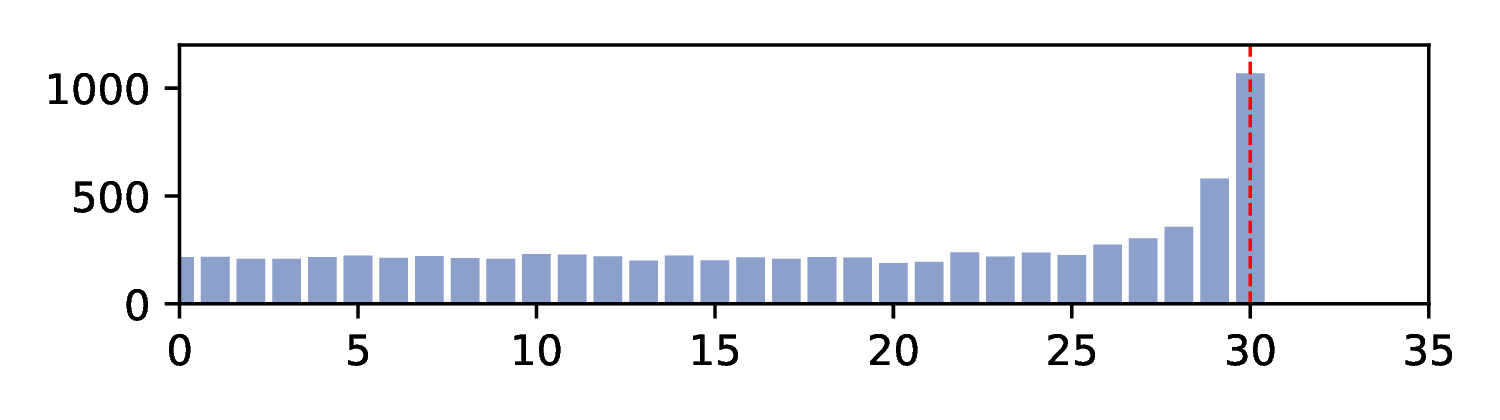} 
                     & \includegraphics[width=0.98\linewidth, trim={168pt 55pt 44pt 26pt}, clip]{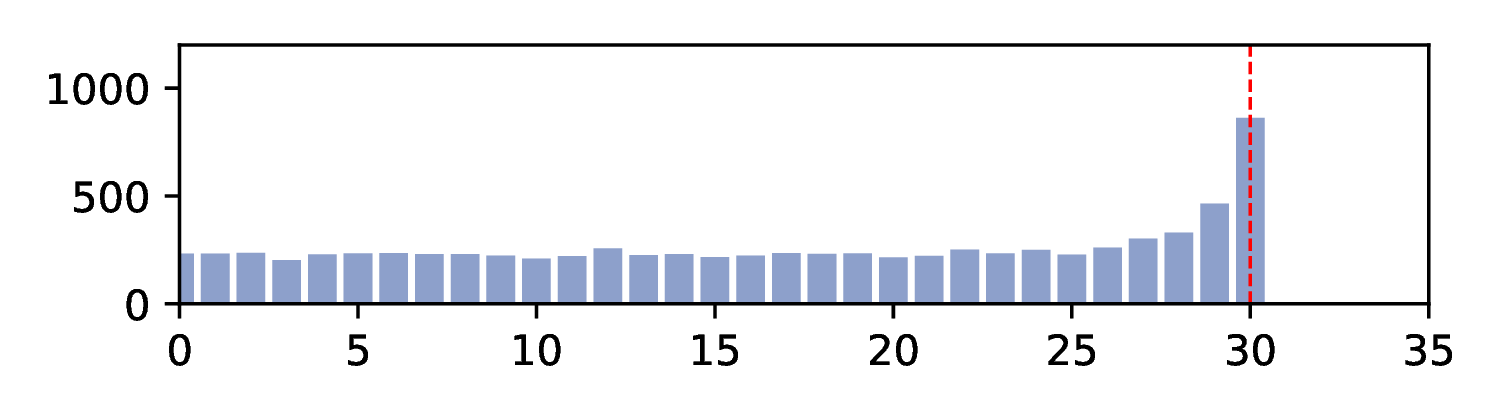}
                     & \includegraphics[width=0.98\linewidth, trim={168pt 55pt 44pt 26pt}, clip]{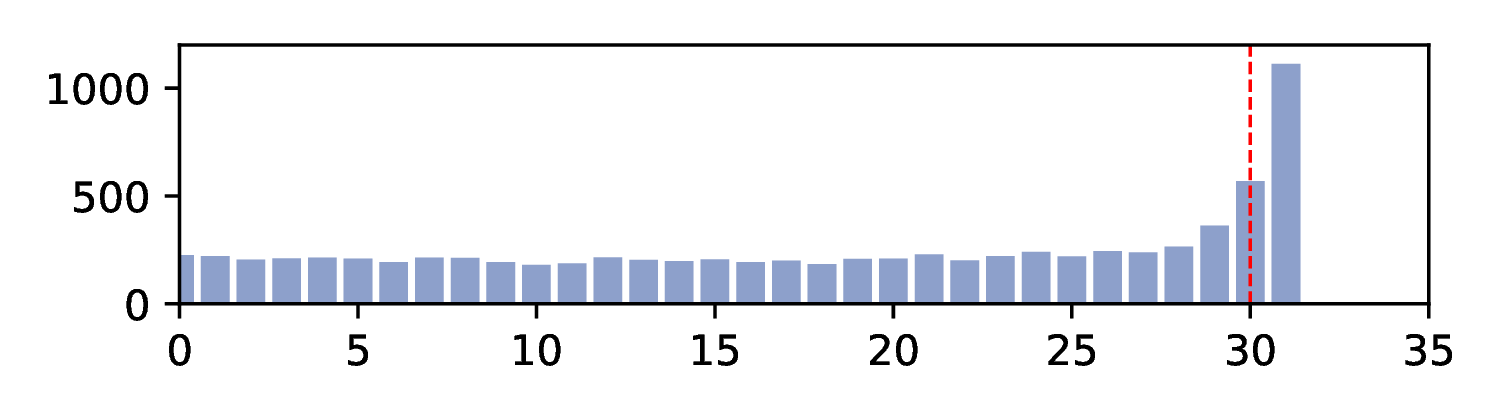} 
                     & \includegraphics[width=0.98\linewidth, trim={168pt 55pt 44pt 26pt}, clip]{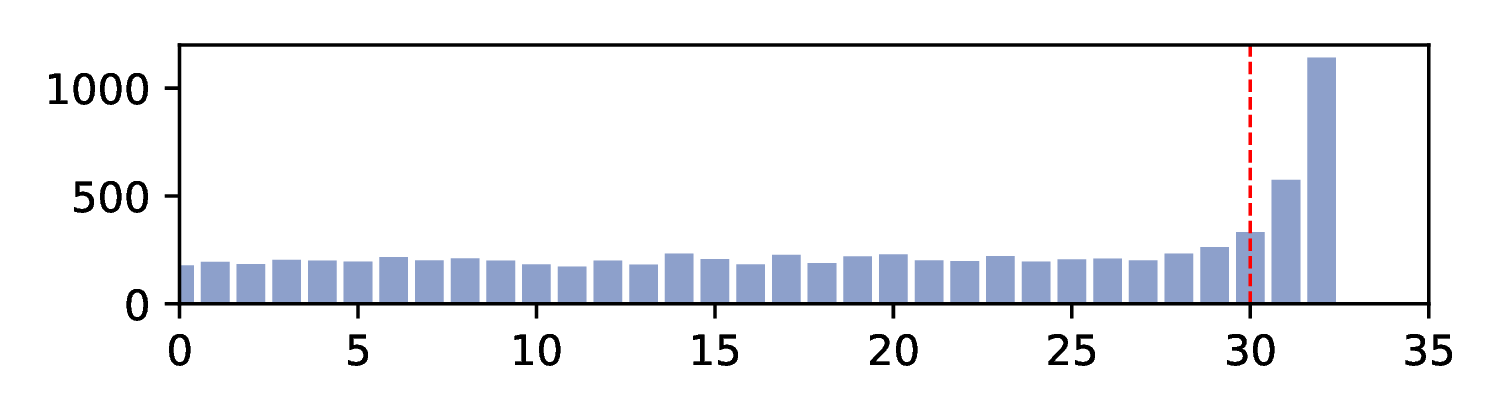}
\\
            & Iteration $S=124$ &  Iteration $S=128$ & Iteration $S=132$ & Iteration $S=141$
        \end{tabularx}%
        
    \makeatletter\def\@captype{figure}\makeatother
    \vspace{-1.5mm}
    \caption{\label{fig:sampling}Sampling count distribution, illustrating the effect of our shifted exponential sampling weight functions with varying $\alpha$ and $\beta$ values at different iterations. The horizontal axis represents the frame index, while the vertical axis indicates the number of sampled rays. Notably, we emphasize the fluctuations in sampling counts for frame \#30, received at iteration 124.}
    \vspace{-0.5cm}
\end{table*}

\subsection{Modeling the Keyframe Stream}
Under the online training setting, each new keyframe $I_i$ is sequentially sent to the NeRF/3DGS training server at time $T_i$. The time interval between two adjacent keyframes $\Delta T_i = T_{i+1} - T_{i}$ varies due to factors like camera or picture motion, system computation, and network fluctuations.
We can loosely model the arrival of new keyframes as a Poisson point process (PPP).
By definition of the Poisson point process, the time interval $\Delta T_i$ between successive keyframes follows an exponential distribution:
\begin{equation}
    \Delta T \sim Exp(\lambda),\quad f(x, \lambda) = \lambda \exp(-\lambda x) \label{eq:exp_distr}
\end{equation}
where $f(x, \lambda)$ is PDF function,  $\lambda$ is the expected rate of arrival of a new keyframe.
This modelling of the keyframe stream gives us insights for designing a new sampling method that emphasizes recent frames while considering the keyframe arrival rate.

\subsection{Sampling with Shifted Exponential Distribution}
\label{sec:exp_sampling}
The sampling problem we address involves determining which keyframes to sample from for each training iteration, given the limited number of total iterations available for online training.
As discussed in Section \ref{sec:intro}, the naive uniform sampling approach does not sufficiently sample the later frames resulting in a rendering quality drop. 
Therefore, we need a sampling method that generates more samples from the more recently received frames during training.

Before presenting our method,  we define how we represent time units for NeRF/3DGS training.
Since sampling is performed once per training iteration, and these iterations take roughly the same amount of time, we use the number of iterations to denote elapsed time. For instance, the timestamp $T_i$ of keyframe $I_i$ corresponds to the $T_i$-th iteration when the training routine first encounters $I_i$.

We seek to define a function $f$ that maps each keyframe to a sampling weight. The earlier frames far behind the current iteration should have lower weights and be sampled more uniformly, while the newer frames should have higher weights.
To ensure this function remains unaffected by the continuous increase in training iterations, we use relative time intervals as input for $f$, instead of absolute timestamps. Specifically, the time interval $D_i$ for a keyframe with timestamp $T_i$ at iteration $S$ is calculated as $D_i = S - T_i$, ensuring that newer frames consistently have lower $D_i$. Thus, the function $f(D)$ should be descending over time.

Inspired by the exponential distribution of time intervals between two adjacent keyframes (Eq. \ref{eq:exp_distr}), we utilize the properties of an exponential distribution to define a sampling weight function\footnote{The sampling weights $W$ will be normalized for the PDF sampling.}:
\begin{equation}
W^*_i = f^*(D_i) = \exp(-\lambda D_i)
\end{equation}
However, simply using this function would severely downweight the earlier frames.
Because of the exponential decrease, the earlier frames will have sampling weights very close to zero and almost all the samples will come from the recent frames. The lack of access to the earlier frames could cause the forgetting issue \cite{french1999catastrophic} in the training which could further cause the overall rendering quality drop~\cite{sucar2021imap, chung2022meil}.

To address this problem, we propose to add an offset term $\beta / N_S$ to ensure the earlier frames still have certain sampling weights to be almost uniformly sampled:
\begin{equation}
W_i = f(D_i) = \exp(-\alpha\lambda D_i) + \beta / N_S
\end{equation}
where $N_S$ is the number of available keyframes at iteration $S$; $\alpha$ and $\beta$ are two hyperparameters:
$\alpha$ scales the average keyframe rate to further control the decrease rate of $f(D)$, $\beta$ can control the ratio of the rays to be sampled from the most recent frames (e.g., $T_{N_S} = S$), because the portion of rays being sampled from frame $I_{N_S}$, given a sufficiently large $N_S$, can be roughly approximated as 
\begin{equation}
    p_{N_S} = \frac{W_{N_S}}{\sum_{i=1}^{N_S}W_i} = \frac{1+\beta / N_S}{\beta + \sum_{i=1}^{N_S} \exp(-\alpha\lambda D_i)} \approx \frac{1}{\beta + 2}
\label{eq:sample_func}
\end{equation}

Fig. \ref{fig:sampling} demonstrates the sampling results with varied $\alpha$ and $\beta$ values. Our proposed sampling function can gradually decrease the sampling weight of a newly received frame as training proceeds. With the help of the $\beta$ term, the sampling for earlier frames is similar to uniform sampling, providing sufficient training samples to avoid the forgetting issue. We empirically find a default setting of $\alpha=2, \beta=4$ would give a decent sampling balance for the online training.

\section{Experiments}
\label{sec:exp}

\subsection{Setup}
To deploy our online reconstruction framework, we use a desktop machine with an RTX4090 GPU as the remote server for reconstruction training. We use a Jetson Orin Nano board as our local edge processor, meeting the SLAM computation needs with low energy consumption.

We deploy a customized ORB-SLAM2 \cite{mur2017orb} as our on-device SLAM module for the local end. 
As mentioned in Sec. \ref{sec:model}, we evaluate our online training with both NeRF and 3DGS as the 3D representation model on the remote server.
We also enable the differentiable pose refinement \cite{wang2021nerf} to make our training more robust to less accurate camera poses from real-time tracking. 

It is challenging to fairly compare different online approaches due to the impact of the randomness of the data generated by the local end and transmitted over the network. 
Therefore, we implement a simulation module on the remote end, replaying the keyframe stream logged by real-time SLAM systems. This approach guarantees that the keyframe stream exposed to the training process remains consistent across different training strategies for an accurate and fair comparison. 
For all the evaluated online training strategies, we ensure that the same number of keyframes is presented across different strategies at any specific training iteration. This is achieved by replaying a keyframe-iteration log, which is recorded by running the na\"ive online training without any
computation interference on the remote end.
The online training stops upon receipt of all keyframes, with only a few additional training iterations (less than 10 seconds). The recorded number of training iterations is then applied across all the evaluated training strategies.
The offline model training in our experiments uses the same set of keyframes but with full access to all the keyframes at the beginning of training.
To quantitatively evaluate the rendering quality, we evaluate scenes with PSNR, SSIM, and LPIPS metrics on the uniformly sampled frames along the camera trajectory.

\begin{table}[]
\vspace{1.5mm}
\caption{Quantitative results on Replica dataset, scores are averaged over 8 scans.}
\label{tab:replica}
\centering
\centering
\small
\footnotesize

\setlength\tabcolsep{0.5pt}
\begin{tabularx}{0.96\linewidth}
    {{>{\arraybackslash}p{4.2em}|}*{3}{>{\centering\arraybackslash}X}{|>{\arraybackslash}p{.1em}|>{\arraybackslash}p{4.2em}|}*{3}{>{\centering\arraybackslash}X}}
    \hline
    \multirow{2}{*}{\textbf{Method}} &  \multicolumn{3}{c|}{\textbf{NeRF}} & &\multirow{2}{*}{\textbf{Method}}  &  \multicolumn{3}{c}{\textbf{3DGS}}
    \\  \hhline{~|---|~|~|---}
     & \textbf{PSNR}  & \textbf{SSIM} & \textbf{LPIPS} & &  & \textbf{PSNR}  & \textbf{SSIM} & \textbf{LPIPS} 
    \\ \hhline{-|---|~|-|---}
    offline & \cellcolor{orange!25}29.63 & \cellcolor{red!25}0.869 & \cellcolor{red!25}0.251            & & offline & \cellcolor{red!25}31.03 & \cellcolor{red!25}0.907 & \cellcolor{red!25}0.173
    \\
    uniform & 28.77 & 0.858 & 0.281                                                                     & & uniform & 29.39 & 0.877 & 0.200
    \\ \hhline{-|---|~|-|---}
    imap & 28.99 & 0.860 & 0.274                                                                        & & imap & 29.07 & 0.875 & 0.205
    \\
    ours & 29.06 & 0.861 & 0.269                                                                         & & ours & \cellcolor{yellow!25}29.89 & 0.882 & 0.190
    \\ \hhline{-|---|~|-|---}
    imap+loss & \cellcolor{yellow!25}29.52 & \cellcolor{yellow!25}0.865 & \cellcolor{yellow!25}0.263    & & imap+mvs & 29.78 & \cellcolor{orange!25}0.883 & \cellcolor{yellow!25}0.187
    \\
    ours+loss & \cellcolor{red!25}29.70 & \cellcolor{red!25}0.869 & \cellcolor{orange!25}0.259           & & ours+mvs & \cellcolor{orange!25}30.01 & \cellcolor{orange!25}0.883 & \cellcolor{orange!25}0.185
    \\ \hline
\end{tabularx}

\vspace{-5mm}
\end{table}

\subsection{Datasets}

We evaluate our method using the Replica~\cite{straub2019replica} and Tanks and Temples~\cite{sturm12iros} datasets. 
Replica contains various synthetic small-scale indoor scenes. We use the same camera trajectory of around 1-minute length from iMAP~\cite{sucar2021imap} for our online NeRF learning. 
The tanks and temples (TnT) dataset has a collection of high-resolution real-captured video recordings (3-7 minutes) of various indoor and outdoor scenes. We pick a small subset of outdoor scenes to showcase the capability of our method for challenging unbounded outdoor scenes.
We utilize ORB-SLAM2 with downscaled frames (640x360) and a reduced frame rate (10 FPS) for more stable tracking performance on the TnT video input.
We use RGB video streams of scenes for all evaluated methods.
We also captured 2 scenes with a handheld compute board, the results are shown in the supplement video\footnote{\url{https://www.youtube.com/watch?v=qphP5QnMp6w}}.

\begin{table}[t]
\vspace{1.5mm}
\caption{Quantitative results on Tanks\&Temples scenes.}
\label{tab:tnt}
\centering
\centering
\small
\footnotesize
\setlength\tabcolsep{0.5pt}
\begin{tabularx}{0.98\linewidth}
    {{>{\centering\arraybackslash}p{1.5em}>{\raggedright\arraybackslash}p{3.5em}|}*{3}{>{\centering\arraybackslash}X}|*{3}{>{\centering\arraybackslash}X}|*{3}{>{\centering\arraybackslash}X}}
\hline
\multicolumn{2}{c|}{\multirow{2}{*}{\textbf{Method}}} &\multicolumn{3}{c|}{Barn} & \multicolumn{3}{c|}{Train} & \multicolumn{3}{c}{Truck}                 
\\ \cline{3-11}
& & \multicolumn{1}{c}{\textbf{\scriptsize PSNR}} & \multicolumn{1}{c}{\textbf{\scriptsize SSIM}} & \multicolumn{1}{c|}{\textbf{\scriptsize LPIPS}} & \multicolumn{1}{c}{\textbf{\scriptsize PSNR}} & \multicolumn{1}{c}{\textbf{\scriptsize SSIM}} & \multicolumn{1}{c|}{\textbf{\scriptsize LPIPS}} & \multicolumn{1}{c}{\textbf{\scriptsize PSNR}} & \multicolumn{1}{c}{\textbf{\scriptsize SSIM}} & \multicolumn{1}{c}{\textbf{\scriptsize LPIPS}} \\ \hline
\multirow{4}{*}{\rotatebox[origin=c]{90}{\textbf{NeRF}}}
& offline               & \cellcolor{red!25}25.30                                       & \cellcolor{red!25}0.721                                       & \cellcolor{red!25}0.290                                           & \cellcolor{red!25}23.54                                       & \cellcolor{red!25}0.725                                       & \cellcolor{red!25}0.232                                           & \cellcolor{red!25}24.78                                       & \cellcolor{red!25}0.791                                       & \cellcolor{red!25}0.168                                          \\
\cline{2-11}
& uniform                & 23.23                                       & 0.627                                       & 0.418                                           & 20.41                                       & 0.554                                       & 0.451                                           & 23.27                                       & 0.723                                       & 0.260                                          \\
& imap         & \cellcolor{yellow!25}23.91                                       & \cellcolor{yellow!25}0.660                                       & \cellcolor{yellow!25}0.368                                           & \cellcolor{yellow!25}21.40                                       & \cellcolor{yellow!25}0.603                                       & \cellcolor{yellow!25}0.378                                           & \cellcolor{yellow!25}23.59                                       & \cellcolor{yellow!25}0.737                                       & \cellcolor{orange!25}0.232                                          \\
& ours         & \cellcolor{orange!25}24.22                                       & \cellcolor{orange!25}0.677                                       & \cellcolor{orange!25}0.345                                           & \cellcolor{orange!25}21.81                                       & \cellcolor{orange!25}0.624                                       & \cellcolor{orange!25}0.352                                           & \cellcolor{orange!25}23.74                                       & \cellcolor{orange!25}0.739                                       & \cellcolor{yellow!25}0.236
\\
\hline \hline
\multirow{4}{*}{\rotatebox[origin=c]{90}{\textbf{3DGS}}}
& offline     & \cellcolor{red!25}26.58                       & \cellcolor{red!25}0.852                       & \cellcolor{red!25}0.142                         & \cellcolor{red!25}21.70                       & \cellcolor{red!25}0.792                       & \cellcolor{red!25}0.178                         & \cellcolor{red!25}24.74                       & \cellcolor{red!25}0.871                       & \cellcolor{red!25}0.109                        \\
\cline{2-11}
& uniform     & 24.22                                         & 0.782                                         & 0.172                                           & 20.46                                         & 0.703                                         & 0.228                                           & 22.26                                         & 0.783                                         & 0.153                                          \\
& imap        & 24.48                                         & 0.798                                         & \cellcolor{yellow!25}0.160                      & 20.44                                         & 0.701                                         & 0.239                                           & 22.47                                         & 0.791                                         & 0.152                                          \\
& ours         & \cellcolor{orange!25}24.67                    & \cellcolor{yellow!25}0.803                    & 0.162                                           & \cellcolor{orange!25}20.73                    & \cellcolor{orange!25}0.722                    & \cellcolor{orange!25}0.217                      & \cellcolor{orange!25}22.75                    & \cellcolor{orange!25}0.803                    & \cellcolor{yellow!25}0.143                     \\
\hline

\end{tabularx}

\vspace{-6mm}
\end{table}

\begin{table*}[t]
    \vspace{1.8mm}
    \centering
    \footnotesize
    \setlength\tabcolsep{0pt}
\begin{tabularx}{\linewidth}
        {>{\centering\arraybackslash}p{1.5em}p{1em}*{6}{>{\centering\arraybackslash}X}}
\hdashline \\[-7pt]
\multirow{3}{*}{\rotatebox[origin=c]{90}{\textbf{NeRF\quad}}}
&\rotatebox[origin=l]{90}{office2} & \multicolumn{6}{c}{\includegraphics[width=0.96\linewidth, trim={0 2cm 0 0},clip]{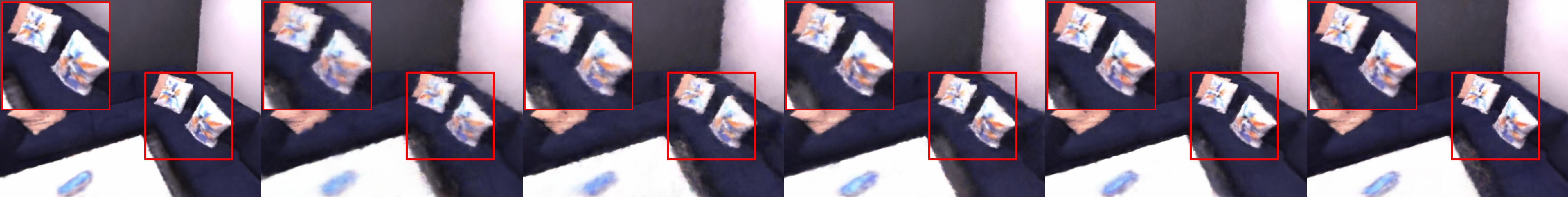}}\\
&\rotatebox[origin=l]{90}{office3} & \multicolumn{6}{c}{\includegraphics[width=0.96\linewidth, trim={0 3cm 0 0},clip]{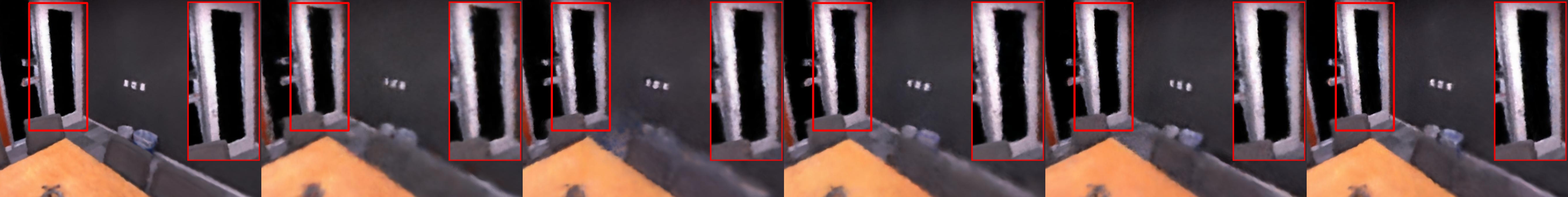}}\\
\hdashline
& & offline & uniform & imap & ours & imap+loss & ours+loss \\[3pt]
\hdashline \\[-7pt]
\multirow{3}{*}{\rotatebox[origin=c]{90}{\textbf{3DGS\quad}}}
&\rotatebox[origin=l]{90}{office0} & \multicolumn{6}{c}{\includegraphics[width=0.96\linewidth, trim={0 1cm 0 0},clip]{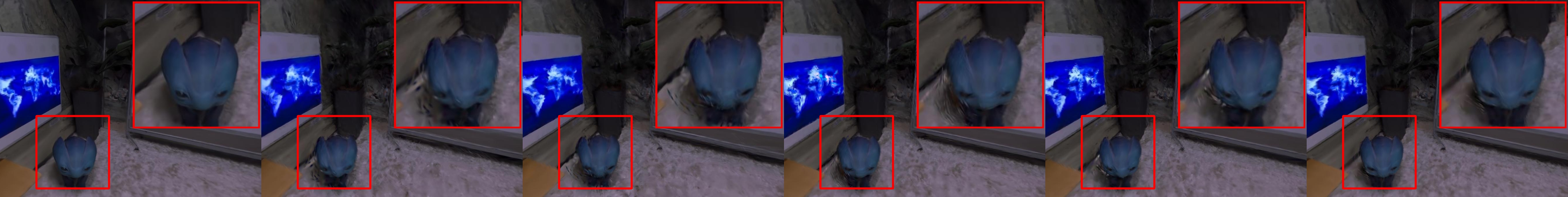}}\\
&\rotatebox[origin=l]{90}{room0} & \multicolumn{6}{c}{\includegraphics[width=0.96\linewidth, trim={0 0 0 0},clip]{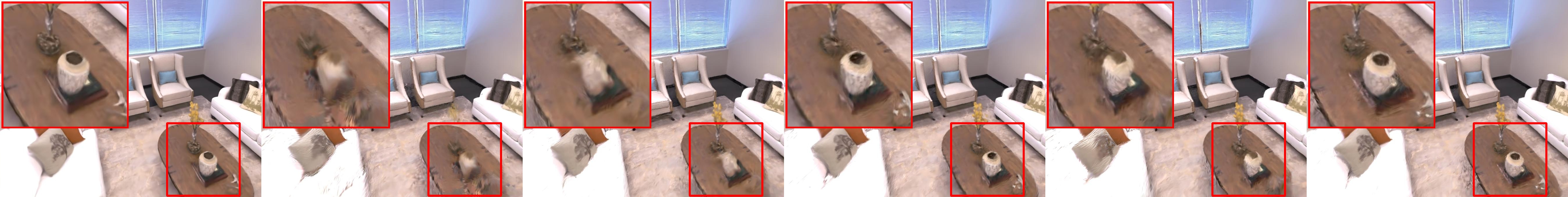}}\\
\hdashline
& & offline & uniform & imap & ours & imap+mvs & ours+mvs \\
\end{tabularx}

\makeatletter\def\@captype{figure}\makeatother
\vspace{-0.3cm}
\caption{Visual comparisons on Replica scenes.
\label{fig:replica}}
\vspace{-4.5mm}
\end{table*}

\begin{table}[]
    \centering
    \setlength\tabcolsep{0pt}
\begin{tabularx}{\linewidth}
        {p{1.3em}p{1em}*{4}{>{\centering\arraybackslash}X}}
\hdashline \\[-7pt]
\multirow{2}{*}{\rotatebox[origin=c]{90}{\textbf{NeRF\quad}}}
& \rotatebox[origin=l]{90}{Barn} & \multicolumn{4}{c}{\includegraphics[width=0.92\linewidth, trim={0 0 0 0},clip]{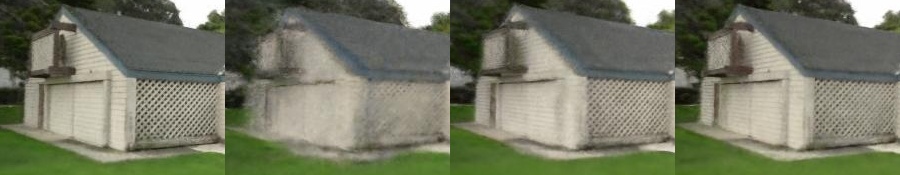}}\\
& \rotatebox[origin=l]{90}{Truck} & \multicolumn{4}{c}{\includegraphics[width=0.92\linewidth, trim={0 5pt 0 16pt},clip]{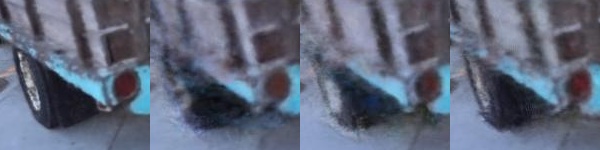}}\\
\hdashline
& & offline & uniform & imap & ours\\[3pt] 
\hdashline \\[-7pt]
\multirow{2}{*}{\rotatebox[origin=c]{90}{\textbf{3DGS\quad}}}
& \rotatebox[origin=l]{90}{Barn} & \multicolumn{4}{c}{\includegraphics[width=0.92\linewidth, trim={0 0 0 0 0},clip]{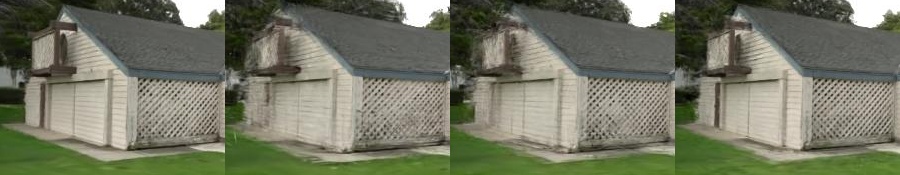}}\\
& \rotatebox[origin=l]{90}{Train} & \multicolumn{4}{c}{\includegraphics[width=0.92\linewidth, trim={0 5pt 0 0 5pt},clip]{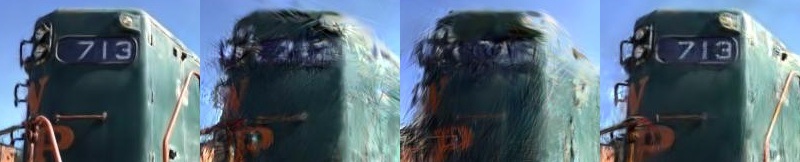}}\\
\hdashline
& & offline & uniform & imap & ours \\
\end{tabularx}

\makeatletter\def\@captype{figure}\makeatother
\vspace{-3mm}
\caption{
Visual comparisons on Tanks\&Temples scenes.
\label{fig:tnt}}
\vspace{-5.5mm}
\end{table}

\subsection{Evaluation}
We first include offline training (``offline") and online training with na\"ive uniform sampling (``uniform") as two fundamental baselines for our comparison.
We then include the methods with different frame sampling strategies. 
We implement a frame sampling method similar to iMAP \cite{sucar2021imap} (``imap frame") for both NeRF and 3DGS. iMAP
always samples $20\%$ rays of the NeRF training batch from the most recent keyframe and uniformly samples the rest of the rays from the remaining keyframes. Since 3DGS training does not involve ray sampling, we instead sample the most recent keyframe with the probability of $20\%$, and $80\%$ for the remaining keyframes.
We use ``ours" to denote our shifted exponential sampling method.
We also combine our proposed frame sampling with other training enhancement methods, to demonstrate its versatility and generalizability.
For NeRF-based online reconstruction, we combine our frame sampling with the loss-guided ray sampling \cite{sucar2021imap, muller2022instant}. This method tracks loss distribution by dividing each image into small patches and maintaining a running average photometric loss for each patch. Rays are then weighted-sampled based on tracked loss. We use ``+loss" to denote methods with loss-guided ray sampling.
For 3DGS-based online reconstruction, we also test our frame sampling with enhanced geometry prior \cite{ fan2024instantsplat}. We leverage DUSt3R \cite{wang2024dust3r}, an efficient state-of-the-art multi-view stereo (MVS) method, to add and initialize Gaussian as estimated geometric priors of the 3D targets. We run DUSt3R on 4 newly received adjacent keyframes ($\sim$3 sec per DUSt3R inference) once every 500 iterations during the first 3000 training iterations to obtain MVS points to be used for adding new Gaussian points. We use ``+mvs" to denote methods enhanced with MVS points.

\subsection{Result Comparison}
\textit{Replica.} 
In Table \ref{tab:replica} and Fig. \ref{fig:replica}, we present quantitative and qualitative results for Replica dataset. These results demonstrate that our shifted exponential sampling method improves the quality of renderings across various scenes. Our frame sampling method (``ours") consistently performs better than the na\"ive uniform sampling and iMAP's frame sampling.
Furthermore, when combined with other training enhancement methods (e.g., ``+loss" and ``+mvs"), our frame sampling can further improve the resulting quality. Our method ``ours+loss" even outperforms the offline NeRF baseline on some scenes. 
The visual results in Fig. \ref{fig:replica} highlight the noticeable quality differences among the different online training methods.
Comparing NeRF and 3DGS, the 3DGS-based methods achieve higher rendering quality. Due to 3DGS's efficient rasterization-based rendering, it can run up to 3 times more training iterations compared to NeRF training in the same timeframe. This efficiency allows 3DGS to reconstruct sharper images with more fine-grained details, particularly for the relatively simple Replica indoor scenes.

\textit{Tanks and Temples.} We evaluate three complex outdoor scenes with results shown in Table \ref{tab:tnt} and Fig. \ref{fig:tnt}. Our frame sampling method still consistently outperforms the na\"ive uniform sampling method and iMAP frame sampling method on most metrics. 
Unlike the results on the Replica dataset, NeRF methods achieve similar or even better quantitative results than 3DGS on these unbounded outdoor scenes. This is because the NeRF method, assisted by spatial contraction as an inductive bias, can better reconstruct 360 unbounded scenes. Due to the lack of reliable geometry priors, reconstructing distant objects in unbounded scenes, especially background objects, is challenging for 3DGS. Therefore, the NeRF-based method is more advantageous for robust reconstruction in our online setup.

\section{Conclusion}
We introduce \model, a distributed framework that enables online 3D reconstruction and visualization of scenes captured by cameras on resource-constrained mobile robots. \model\ leverages posed keyframes computed by on-device SLAM and the computational power of a remote server to facilitate online NeRF/3DGS training.
One key challenge we observed with na\"ive online training compared to offline training is the unbalanced frame sampling during training, leading to a significant loss in quality. To address this challenge, we propose a shifted exponential frame sampling method that adaptively gives more weight to more recent frames while still sufficiently sampling earlier frames.
Experimental results on various datasets and models demonstrate the effectiveness and applicability of our proposed sampling method. We believe our modular framework can enable more real-time use cases and downstream robotics tasks by leveraging high-quality online visualization and 3D representation.

{\small
\bibliographystyle{IEEEtran}
\bibliography{ref}

\begin{thebibliography}{10}
\providecommand{\url}[1]{#1}
\csname url@rmstyle\endcsname
\providecommand{\newblock}{\relax}
\providecommand{\bibinfo}[2]{#2}
\providecommand\BIBentrySTDinterwordspacing{\spaceskip=0pt\relax}
\providecommand\BIBentryALTinterwordstretchfactor{4}
\providecommand\BIBentryALTinterwordspacing{\spaceskip=\fontdimen2\font plus
\BIBentryALTinterwordstretchfactor\fontdimen3\font minus
  \fontdimen4\font\relax}
\providecommand\BIBforeignlanguage[2]{{%
\expandafter\ifx\csname l@#1\endcsname\relax
\typeout{** WARNING: IEEEtran.bst: No hyphenation pattern has been}%
\typeout{** loaded for the language `#1'. Using the pattern for}%
\typeout{** the default language instead.}%
\else
\language=\csname l@#1\endcsname
\fi
#2}}

\bibitem{voxel_hashing}
M.~Nießner, M.~Zollhöfer, S.~Izadi, and M.~Stamminger, ``Real-time 3d
  reconstruction at scale using voxel hashing,'' \emph{ACM Transactions on
  Graphics (TOG)}, vol.~32, 11 2013.

\bibitem{bundlefusion}
A.~Dai, M.~Nie{\ss}ner, M.~Zollh{\"o}fer, S.~Izadi, and C.~Theobalt,
  ``Bundlefusion: Real-time globally consistent 3d reconstruction using
  on-the-fly surface reintegration,'' \emph{ACM Transactions on Graphics
  (ToG)}, vol.~36, no.~4, p.~1, 2017.

\bibitem{whelan2016elasticfusion}
T.~Whelan, R.~F. Salas-Moreno, B.~Glocker, A.~J. Davison, and S.~Leutenegger,
  ``Elasticfusion: Real-time dense slam and light source estimation,''
  \emph{The International Journal of Robotics Research}, vol.~35, no.~14, pp.
  1697--1716, 2016.

\bibitem{mildenhall2020nerf}
B.~Mildenhall, P.~P. Srinivasan, M.~Tancik, J.~T. Barron, R.~Ramamoorthi, and
  R.~Ng, ``Nerf: Representing scenes as neural radiance fields for view
  synthesis,'' in \emph{European Conference on Computer Vision}.\hskip 1em plus
  0.5em minus 0.4em\relax Springer, 2020, pp. 405--421.

\bibitem{kerbl20233d}
B.~Kerbl, G.~Kopanas, T.~Leimk{\"u}hler, and G.~Drettakis, ``3d gaussian
  splatting for real-time radiance field rendering,'' \emph{ACM Transactions on
  Graphics (ToG)}, vol.~42, no.~4, pp. 1--14, 2023.

\bibitem{muller2022instant}
T.~M{\"u}ller, A.~Evans, C.~Schied, and A.~Keller, ``Instant neural graphics
  primitives with a multiresolution hash encoding,'' \emph{ACM Transactions on
  Graphics (ToG)}, vol.~41, no.~4, pp. 1--15, 2022.

\bibitem{instant3d}
S.~Li, C.~Li, W.~Zhu, B.~Yu, Y.~Zhao, C.~Wan, H.~You, H.~Shi, and Y.~Lin,
  ``Instant-3d: Instant neural radiance field training towards on-device ar/vr
  3d reconstruction,'' in \emph{Proceedings of the International Symposium on
  Computer Architecture}, 2023, pp. 1--13.

\bibitem{yu2023nerfbridge}
J.~Yu, J.~E. Low, K.~Nagami, and M.~Schwager, ``Nerfbridge: Bringing real-time,
  online neural radiance field training to robotics,'' \emph{arXiv preprint
  arXiv:2305.09761}, 2023.

\bibitem{mur2017orb}
R.~Mur-Artal and J.~D. Tard{\'o}s, ``Orb-slam2: An open-source slam system for
  monocular, stereo, and rgb-d cameras,'' \emph{IEEE transactions on robotics},
  vol.~33, no.~5, pp. 1255--1262, 2017.

\bibitem{schoenberger2016sfm}
J.~L. Sch\"{o}nberger and J.-M. Frahm, ``Structure-from-motion revisited,'' in
  \emph{Conference on Computer Vision and Pattern Recognition (CVPR)}, 2016.

\bibitem{6162880}
R.~A. Newcombe, S.~Izadi, O.~Hilliges, D.~Molyneaux, D.~Kim, A.~J. Davison,
  P.~Kohi, J.~Shotton, S.~Hodges, and A.~Fitzgibbon, ``Kinectfusion: Real-time
  dense surface mapping and tracking,'' in \emph{2011 10th IEEE International
  Symposium on Mixed and Augmented Reality}, 2011, pp. 127--136.

\bibitem{keller2013real}
M.~Keller, D.~Lefloch, M.~Lambers, S.~Izadi, T.~Weyrich, and A.~Kolb,
  ``Real-time 3d reconstruction in dynamic scenes using point-based fusion,''
  in \emph{2013 International Conference on 3D Vision-3DV 2013}.\hskip 1em plus
  0.5em minus 0.4em\relax IEEE, 2013, pp. 1--8.

\bibitem{cao2018real}
Y.-P. Cao, L.~Kobbelt, and S.-M. Hu, ``Real-time high-accuracy
  three-dimensional reconstruction with consumer rgb-d cameras,'' \emph{ACM
  Transactions on Graphics (TOG)}, vol.~37, no.~5, pp. 1--16, 2018.

\bibitem{sucar2020nodeslam}
E.~Sucar, K.~Wada, and A.~Davison, ``Nodeslam: Neural object descriptors for
  multi-view shape reconstruction,'' in \emph{2020 International Conference on
  3D Vision (3DV)}.\hskip 1em plus 0.5em minus 0.4em\relax IEEE, 2020, pp.
  949--958.

\bibitem{weder2021neuralfusion}
S.~Weder, J.~L. Schonberger, M.~Pollefeys, and M.~R. Oswald, ``Neuralfusion:
  Online depth fusion in latent space,'' in \emph{Proceedings of the IEEE/CVF
  Conference on Computer Vision and Pattern Recognition}, 2021, pp. 3162--3172.

\bibitem{sun2021neuralrecon}
J.~Sun, Y.~Xie, L.~Chen, X.~Zhou, and H.~Bao, ``Neuralrecon: Real-time coherent
  3d reconstruction from monocular video,'' in \emph{Proceedings of the
  IEEE/CVF Conference on Computer Vision and Pattern Recognition}, 2021, pp.
  15\,598--15\,607.

\bibitem{teed2021droid}
Z.~Teed and J.~Deng, ``Droid-slam: Deep visual slam for monocular, stereo, and
  rgb-d cameras,'' \emph{Advances in neural information processing systems},
  vol.~34, pp. 16\,558--16\,569, 2021.

\bibitem{chen2022tensorf}
A.~Chen, Z.~Xu, A.~Geiger, J.~Yu, and H.~Su, ``Tensorf: Tensorial radiance
  fields,'' in \emph{European Conference on Computer Vision}.\hskip 1em plus
  0.5em minus 0.4em\relax Springer, 2022, pp. 333--350.

\bibitem{kerr2022evo}
J.~Kerr, L.~Fu, H.~Huang, Y.~Avigal, M.~Tancik, J.~Ichnowski, A.~Kanazawa, and
  K.~Goldberg, ``Evo-nerf: Evolving nerf for sequential robot grasping of
  transparent objects,'' in \emph{6th Annual Conference on Robot Learning},
  2022.

\bibitem{yen2022nerf}
L.~Yen-Chen, P.~Florence, J.~T. Barron, T.-Y. Lin, A.~Rodriguez, and P.~Isola,
  ``Nerf-supervision: Learning dense object descriptors from neural radiance
  fields,'' in \emph{2022 International Conference on Robotics and Automation
  (ICRA)}.\hskip 1em plus 0.5em minus 0.4em\relax IEEE, 2022, pp. 6496--6503.

\bibitem{byravan2023nerf2real}
A.~Byravan, J.~Humplik, L.~Hasenclever, A.~Brussee, F.~Nori, T.~Haarnoja,
  B.~Moran, S.~Bohez, F.~Sadeghi, B.~Vujatovic, \emph{et~al.}, ``Nerf2real:
  Sim2real transfer of vision-guided bipedal motion skills using neural
  radiance fields,'' in \emph{2023 IEEE International Conference on Robotics
  and Automation (ICRA)}.\hskip 1em plus 0.5em minus 0.4em\relax IEEE, 2023,
  pp. 9362--9369.

\bibitem{adamkiewicz2022vision}
M.~Adamkiewicz, T.~Chen, A.~Caccavale, R.~Gardner, P.~Culbertson, J.~Bohg, and
  M.~Schwager, ``Vision-only robot navigation in a neural radiance world,''
  \emph{IEEE Robotics and Automation Letters}, vol.~7, no.~2, pp. 4606--4613,
  2022.

\bibitem{lu2024manigaussian}
G.~Lu, S.~Zhang, Z.~Wang, C.~Liu, J.~Lu, and Y.~Tang, ``Manigaussian: Dynamic
  gaussian splatting for multi-task robotic manipulation,'' \emph{arXiv
  preprint arXiv:2403.08321}, 2024.

\bibitem{wang2024query}
J.~Wang, Z.~Zhang, Q.~Zhang, J.~Li, J.~Sun, M.~Sun, J.~He, and R.~Xu,
  ``Query-based semantic gaussian field for scene representation in
  reinforcement learning,'' \emph{arXiv preprint arXiv:2406.02370}, 2024.

\bibitem{cai2023clnerf}
Z.~Cai and M.~M{\"u}ller, ``Clnerf: Continual learning meets nerf,'' in
  \emph{Proceedings of the IEEE/CVF International Conference on Computer
  Vision}, 2023, pp. 23\,185--23\,194.

\bibitem{rebuffi2017icarl}
S.-A. Rebuffi, A.~Kolesnikov, G.~Sperl, and C.~H. Lampert, ``icarl: Incremental
  classifier and representation learning,'' in \emph{Proceedings of the IEEE
  conference on Computer Vision and Pattern Recognition}, 2017, pp. 2001--2010.

\bibitem{sucar2021imap}
E.~Sucar, S.~Liu, J.~Ortiz, and A.~J. Davison, ``imap: Implicit mapping and
  positioning in real-time,'' in \emph{Proceedings of the IEEE/CVF
  International Conference on Computer Vision}, 2021, pp. 6229--6238.

\bibitem{zhu2022nice}
Z.~Zhu, S.~Peng, V.~Larsson, W.~Xu, H.~Bao, Z.~Cui, M.~R. Oswald, and
  M.~Pollefeys, ``Nice-slam: Neural implicit scalable encoding for slam,'' in
  \emph{Proceedings of the IEEE/CVF Conference on Computer Vision and Pattern
  Recognition}, 2022, pp. 12\,786--12\,796.

\bibitem{matsuki2024gaussian}
H.~Matsuki, R.~Murai, P.~H. Kelly, and A.~J. Davison, ``Gaussian splatting
  slam,'' in \emph{Proceedings of the IEEE/CVF Conference on Computer Vision
  and Pattern Recognition}, 2024, pp. 18\,039--18\,048.

\bibitem{keetha2024splatam}
N.~Keetha, J.~Karhade, K.~M. Jatavallabhula, G.~Yang, S.~Scherer, D.~Ramanan,
  and J.~Luiten, ``Splatam: Splat track \& map 3d gaussians for dense rgb-d
  slam,'' in \emph{Proceedings of the IEEE/CVF Conference on Computer Vision
  and Pattern Recognition}, 2024, pp. 21\,357--21\,366.

\bibitem{huang2024photo}
H.~Huang, L.~Li, H.~Cheng, and S.-K. Yeung, ``Photo-slam: Real-time
  simultaneous localization and photorealistic mapping for monocular stereo and
  rgb-d cameras,'' in \emph{Proceedings of the IEEE/CVF Conference on Computer
  Vision and Pattern Recognition}, 2024, pp. 21\,584--21\,593.

\bibitem{qin2018vins}
T.~Qin, P.~Li, and S.~Shen, ``Vins-mono: A robust and versatile monocular
  visual-inertial state estimator,'' \emph{IEEE transactions on robotics},
  vol.~34, no.~4, pp. 1004--1020, 2018.

\bibitem{labbe2019rtab}
M.~Labb{\'e} and F.~Michaud, ``Rtab-map as an open-source lidar and visual
  simultaneous localization and mapping library for large-scale and long-term
  online operation,'' \emph{Journal of field robotics}, vol.~36, no.~2, pp.
  416--446, 2019.

\bibitem{tancik2023nerfstudio}
M.~Tancik, E.~Weber, E.~Ng, R.~Li, B.~Yi, T.~Wang, A.~Kristoffersen, J.~Austin,
  K.~Salahi, A.~Ahuja, \emph{et~al.}, ``Nerfstudio: A modular framework for
  neural radiance field development,'' in \emph{ACM SIGGRAPH 2023 Conference
  Proceedings}, 2023, pp. 1--12.

\bibitem{quigley2009ros}
M.~Quigley, K.~Conley, B.~Gerkey, J.~Faust, T.~Foote, J.~Leibs, R.~Wheeler,
  A.~Y. Ng, \emph{et~al.}, ``Ros: an open-source robot operating system,'' in
  \emph{ICRA workshop on open source software}, vol.~3, no. 3.2.\hskip 1em plus
  0.5em minus 0.4em\relax Kobe, Japan, 2009, p.~5.

\bibitem{barron2022mip}
J.~T. Barron, B.~Mildenhall, D.~Verbin, P.~P. Srinivasan, and P.~Hedman,
  ``Mip-nerf 360: Unbounded anti-aliased neural radiance fields,'' in
  \emph{Proceedings of the IEEE/CVF Conference on Computer Vision and Pattern
  Recognition}, 2022, pp. 5470--5479.

\bibitem{jung2024relaxing}
J.~Jung, J.~Han, H.~An, J.~Kang, S.~Park, and S.~Kim, ``Relaxing accurate
  initialization constraint for 3d gaussian splatting,'' \emph{arXiv preprint
  arXiv:2403.09413}, 2024.

\bibitem{french1999catastrophic}
R.~M. French, ``Catastrophic forgetting in connectionist networks,''
  \emph{Trends in cognitive sciences}, vol.~3, no.~4, pp. 128--135, 1999.

\bibitem{chung2022meil}
J.~Chung, K.~Lee, S.~Baik, and K.~M. Lee, ``Meil-nerf: Memory-efficient
  incremental learning of neural radiance fields,'' \emph{arXiv preprint
  arXiv:2212.08328}, 2022.

\bibitem{wang2021nerf}
Z.~Wang, S.~Wu, W.~Xie, M.~Chen, and V.~A. Prisacariu, ``Nerf--: Neural
  radiance fields without known camera parameters,'' \emph{arXiv preprint
  arXiv:2102.07064}, 2021.

\bibitem{straub2019replica}
J.~Straub, T.~Whelan, L.~Ma, Y.~Chen, E.~Wijmans, S.~Green, J.~J. Engel,
  R.~Mur-Artal, C.~Ren, S.~Verma, \emph{et~al.}, ``The replica dataset: A
  digital replica of indoor spaces,'' \emph{arXiv preprint arXiv:1906.05797},
  2019.

\bibitem{sturm12iros}
J.~Sturm, N.~Engelhard, F.~Endres, W.~Burgard, and D.~Cremers, ``A benchmark
  for the evaluation of rgb-d slam systems,'' in \emph{Proc. of the
  International Conference on Intelligent Robot Systems (IROS)}, Oct. 2012.

\bibitem{fan2024instantsplat}
Z.~Fan, W.~Cong, K.~Wen, K.~Wang, J.~Zhang, X.~Ding, D.~Xu, B.~Ivanovic,
  M.~Pavone, G.~Pavlakos, \emph{et~al.}, ``Instantsplat: Unbounded sparse-view
  pose-free gaussian splatting in 40 seconds,'' \emph{arXiv preprint
  arXiv:2403.20309}, 2024.

\bibitem{wang2024dust3r}
S.~Wang, V.~Leroy, Y.~Cabon, B.~Chidlovskii, and J.~Revaud, ``Dust3r: Geometric
  3d vision made easy,'' in \emph{Proceedings of the IEEE/CVF Conference on
  Computer Vision and Pattern Recognition}, 2024, pp. 20\,697--20\,709.

\end{thebibliography}
}

\end{document}